\documentclass[10pt,journal,compsoc]{IEEEtran}
\usepackage{latexsym}
\usepackage{amsfonts,amsmath}
\usepackage{url}
\usepackage{paralist}
\usepackage{multirow}
\usepackage{multicol}
\usepackage{caption}
\usepackage{enumerate}
\usepackage{graphicx}
\usepackage{enumitem}
\usepackage{bm}
\usepackage{url}
\usepackage[normalem]{ulem}
\usepackage{subfigure}
\usepackage{mathrsfs}

\usepackage[utf8]{inputenc}
\usepackage{graphicx}
\usepackage{subfigure}
\usepackage{multirow}

\usepackage{microtype}

\usepackage{bm}
\usepackage{booktabs}
\usepackage{bbding}

\usepackage{enumerate}
\usepackage{enumitem}

\usepackage[dvipsnames]{xcolor}

\usepackage[nameinlink]{cleveref}
\crefformat{section}{\S#2#1#3} % see manual of cleveref, section 8.2.1
\crefname{algorithm}{Alg.}{Algs.}
\crefformat{subsection}{\S#2#1#3}
\Crefname{equation}{Eq.}{Eqs.}
\Crefname{figure}{Fig.}{Figs.}

\ifCLASSOPTIONcompsoc
  \usepackage[nocompress]{cite}
\else
  \usepackage{cite}
\fi

\ifCLASSINFOpdf
\else
\fi
\begin{document}
\title{Improving Conversational Recommender System \\ via Contextual and Time-Aware Modeling with \\ Less Domain-Specific Knowledge}

\author{Lingzhi Wang, Shafiq Joty, Wei Gao, Xingshan Zeng, Kam-Fai Wong % <-this % stops a space
\IEEEcompsocitemizethanks{
\IEEEcompsocthanksitem Lingzhi Wang and Kam-Fai Wong are with The Chinese University of Hong Kong, Hong Kong.\protect\\
E-mail: lzwang@se.cuhk.edu.hk, kfwong@se.cuhk.edu.hk

\IEEEcompsocthanksitem Shafiq Joty is with Nanyang Technological University, Singapore.\protect\\
E-mail: srjoty@ntu.edu.sg

\IEEEcompsocthanksitem Wei Gao is with Singapore Management University, Singapore.\protect\\
E-mail: weigao@smu.edu.sg

\IEEEcompsocthanksitem Xingshan Zeng is with Huawei Noah’s Ark Lab, Hong Kong.\protect\\
E-mail: zeng.xingshan@huawei.com
}

\thanks{Manuscript received April 19, 2005; revised August 26, 2015.}}


\markboth{Journal of \LaTeX\ Class Files,~Vol.~14, No.~8, August~2015}%
{Shell \MakeLowercase{\textit{et al.}}: Bare Demo of IEEEtran.cls for Computer Society Journals}



\newcommand{\mlp}{\mathop{\mathrm{MLP}}}
\newcommand{\relu}{\mathop{\mathrm{ReLU}}}
\newcommand{\mask}{\mathop{\mathrm{mask}}}
\newcommand{\argmax}{\mathop{\mathrm{argmax}}}
\newcommand{\softmax}{\mathop{\mathrm{softmax}}}
\newcommand{\enc}{\mathop{\mathrm{Transformer\_Encoder}}}
\newcommand{\dec}{\mathop{\mathrm{Transformer\_Decoder}}}
\newcommand{\ffn}{\mathop{\mathrm{FFN}}}
\newcommand{\selfatt}{\mathop{\mathrm{Self\_Attention}}}
\newcommand{\embed}{\mathop{\mathrm{Embed}}}

\IEEEtitleabstractindextext{%
\begin{abstract}
Conversational Recommender Systems (CRS) has become an emerging research topic seeking to perform recommendations through interactive conversations, which generally consist of generation and recommendation modules. Prior work on CRS tends to incorporate more external and domain-specific knowledge like item reviews to enhance performance. Despite the fact that the collection and annotation of the \textbf{external domain-specific} information needs much human effort and degenerates the generalizability, too much extra knowledge introduces more difficulty to balance among them. Therefore, we propose to fully discover and extract the \textbf{internal} knowledge from the context. We capture both entity-level and contextual-level representations to jointly model user preferences for the recommendation, where a time-aware attention is designed to emphasize the recently appeared items in entity-level representations. We further use the pre-trained BART to initialize the generation module to alleviate the data scarcity and enhance the context modeling. In addition to conducting experiments on a popular dataset (ReDial), we also include a multi-domain dataset (OpenDialKG) to show the effectiveness of our model. Experiments on both datasets show that our model achieves better performance on most evaluation metrics with less external knowledge and generalizes well to other domains. Additional analyses on the recommendation and generation tasks demonstrate the effectiveness of our model in different scenarios.
\end{abstract}

\begin{IEEEkeywords}
Recommender System, Conversational Recommendation, Pre-trained Language Model.
\end{IEEEkeywords}}

\maketitle

\IEEEdisplaynontitleabstractindextext

% \fi
\IEEEpeerreviewmaketitle

\section{Introduction}
Conversational Recommender Systems (\textbf{CRS}) \cite{li2018towards,chen-etal-2019-towards,zhou2020improving,lu2021revcore} have recently attracted many researchers due to the booming of e-commerce platforms. A CRS aims to provide high-quality recommendations to users through conversations. Different from the traditional recommender systems, it focuses on learning users' preferences through natural language interaction with users, and has a high impact in e-commerce. 

An effective CRS is expected to be able to clarify user intents, learn user preferences, recommend high-quality items and reply to users with suitable responses. As we can see from \Cref{tab:intro_case}, the recommender (CRS) recommends items (movies, books, etc.) to the seeker (user) by using natural and fluent sentences. Therefore, most of the studies on the CRS field generally divide the system into two parts: a \emph{recommendation} module to recommend items with top probabilities and a \emph{generation} module to generate responses containing the recommended items. 

For the recommendation module, previous works \cite{zhou2020improving,lu2021revcore} tend to include more and more external knowledge into the system, as most of the available CRS datasets are relatively small (due to the expensive annotation process)~\cite{li2018towards,moon2019opendialkg} and hard to extract meaningful features based on the context alone. For example, to improve the performance in conversational movie recommendation, external knowledge like entity-level knowledge graph \cite{chen-etal-2019-towards}, word-level knowledge graph \cite{zhou2020improving} and item reviews \cite{lu2021revcore} is successively introduced into the system. However, there are three issues existing in the development of including more external knowledge. \textbf{(i)} Though the performance is improved by introducing more external knowledge, how to balance them in a single system becomes a new challenge. \textbf{(ii)} The collection and annotation of external knowledge need much human effort. \textbf{(iii)} Some of the collected external knowledge may be domain-specific and lack generalizability when facing broader application scenarios, e.g., the item reviews introduced by \cite{lu2021revcore} (some recommended items like sports or unpopular items might lack reviews). On the other hand, the previous work largely ignores the timing or sequence order information by regarding the appeared items in context as a set rather than a sequence, which is contrary to the goal of the system. As one of the goals of CRS is to guide the users to express their preferences explicitly through conversations, we believe that the timing or sequence information would be a crucial factor in further improving the performance of a CRS.

Therefore, instead of exploring more domain-specific external knowledge to assist the learning of user preferences, we choose to fully discover and extract the internal knowledge from the context and propose a time-aware user preference modeling. Concretely, we capture both entity-level and contextual-level representations to model the user preferences. The entity-level representations summarize user preferences with the appeared items in context by our proposed time-aware modeling. As for the contextual-level representations extracted by a context encoder, they reflect the semantic- and discourse-level user preferences, which cannot be captured by entity-level ones. 
These two representations complement each other to enhance the recommendation results.

To elaborate the motivation of our recommendation module design, Table \ref{tab:intro_case} shows two conversation examples from CRS datasets, where the upper conversation focuses on recommending movies and the lower one is about recommending books. It is observed that the context contains rich and key information. For example, ``my kids \underline{don't like} \textit{Power Rangers (2017)}'' in the first example of Table \ref{tab:intro_case} indicates the user shows negative opinion to \textit{Power Rangers (2017)} and the CRS is supposed to not recommend items that are similar to the negative item. Only focusing on the appeared items (e.g., \textit{Power Rangers (2017)}) like the previous work \cite{li2018towards,chen-etal-2019-towards,zhou2020improving,lu2021revcore} would misunderstand the users' true preference. In addition to the potential misunderstanding of the user intention, there is also rich and direct information that is vital for a recommendation. For example, ``Do you know any other books \underline{she} wrote.'' in the second example of Table \ref{tab:intro_case} implies that the user needs a specific book recommendation from the aforementioned author \textit{Virginia Woolf}. Such kind of context information provides precise user preference, which should not be ignored by a CRS. Therefore, in addition to the entity-level representation, it's essential to consider the contextual-level user interest representation. 

\begin{table}[t]
\definecolor{mcolor1}{RGB}{41,18,171}
\caption{Two conversation examples from ReDial (the upper row) and OpenDialKG (the lower row) datasets. \textbf{Seeker} is a user who asks movie (or book) recommendation and \textbf{Recommender} is supposed to do chit-chat and recommendation. The mentioned items (i.e., movie and book) are in \textcolor{mcolor1}{blue}.
    }
    % \vskip -1em
    \centering
    \resizebox{\linewidth}{!}{\begin{tabular}{p{7.2cm}}
    % \hline
    \toprule[1.2pt]
    \textbf{\textit{Seeker:}} Can you recommend some newer released family friendly movies? \\
    \textbf{\textit{Recommender:}} Yes you should watch \textcolor{mcolor1}{Power Rangers  (2017)}, \textcolor{mcolor1}{Captain Underpants: The First Epic Movie} was great too \\
    \textbf{\textit{Seeker:}} Well that is a good suggestion, but my kids don't like \textcolor{mcolor1}{Power Rangers  (2017)} much.\\
    \textbf{\textit{Recommender:}} How old are they? \\
    \textbf{\textit{Seeker:}} They are 14 years old and 10 years old. \\
    \textbf{\textit{Recommender:}} \textcolor{mcolor1}{Despicable Me (2010)}  and \textcolor{mcolor1}{Wonder Woman  (2017)} are great flicks \\
    \midrule[1.0pt]
    \textbf{\textit{Seeker:}} Can you recommend a good book by Virginia Woolf? \\
    \textbf{\textit{Recommender:}} Sure! Have your read the \textcolor{mcolor1}{To The Lighthouse}? \\
    \textbf{\textit{Seeker:}} No I have not.  When was it released? \\
    \textbf{\textit{Recommender:}} It came out in 1927. \\
    \textbf{\textit{Seeker:}} Do you know any other books she wrote? \\
    \textbf{\textit{Recommender:}} She wrote many.  You may be interested in \textcolor{mcolor1}{The Waves} or \textcolor{mcolor1}{Mrs. Dalloway}. \\
    \bottomrule[1.2pt]
    \end{tabular}}
    % \vskip -1em
    
    % \vskip -1em
    \label{tab:intro_case}
\end{table}
As for the generation module, most of the previous methods \cite{li2018towards,chen-etal-2019-towards,zhou2020improving,lu2021revcore} employ a general encoder-decoder framework and train the model from scratch. It suffers from the overfitting issue on the relatively small dataset, as most of the CRS datasets only contain about 10k conversations and most of the syntactic structure of the utterances is simple and boring. Therefore, we use the pre-trained BART~\cite{lewis-etal-2020-bart} model to initialize our generation module and thus alleviate the data scarcity effect in capturing meaningful context features. 

Different from most previous works \cite{li2018towards,chen-etal-2019-towards,zhou2020improving,lu2021revcore,wang2021finetuning} that only test the effectiveness of their models on a single domain data, we conduct experiments on two public and popular CRS datasets ReDial~\cite{li2018towards} and OpenDialKG~\cite{moon2019opendialkg} under the multi-domain scenario. The results show that our model can achieve better recommendation and generation performance when assisted with less domain-specific external knowledge, and validate that our model generalizes well to other domains. Further analyses also demonstrate the effectiveness of our proposed modules.

The main contributions of this work can be summarized as follows:

\begin{itemize}
\item We propose to combine both entity-level and contextual-level representations for modeling user preference in conversational recommendation, which achieves better performance with less domain-specific external knowledge compared to the previous works and generalizes well to other domains.

\item We point out the limitation of the previous entity modeling method and contribute a time-aware user preference modeling method to enhance the recommendation.

\item We adopt the pre-trained language model BART to enhance the semantic learning and diversity of our generated responses.
\end{itemize}

The remainder of this paper is organized as follows. The related work is surveyed in Section \ref{sec:relatedwork}.
Section \ref{sec:method} presents the proposed approach. And
Section \ref{sec:exp_setup} and Section \ref{sec:exp_res} present the experimental setup and the corresponding results respectively. Finally, conclusions are drawn in Section \ref{sec:conlusion}.

\section{Related Work}\label{sec:relatedwork}
In this section, we provide an in-depth review of the related research work from three different aspects, conversational recommender system, traditional recommender system and pre-trained languages model. 

\subsection{Conversational Recommender System}
Conversational Recommender System (CRS) has attracted many researchers' interest in recent years. Various task formulations with different hypotheses and application scenarios have been proposed. We summarize them into three categories and introduce them in detail as below. 

\subsubsection{Question Driven Systems}
As the rating or click feedback in traditional recommender systems is limited in that they do not exactly tell why the users like or dislike an item, the feedback from the users could be very sparse. Therefore, question driven systems are proposed to effectively understand users' preferences and improve the recommendations over time by asking clarifying questions. \cite{christakopoulou2018q} proposes question-based video recommender system. \cite{zhang2018towards} builds systems based on aspect-centered questions. \cite{aliannejadi2019asking} formulates the task of asking clarifying questions in open-domain information-seeking conversational systems. More recent works focus on asking attribute-central questions and developing reinforcement learning based approaches \cite{lei2020estimation,ren2020crsal,deng2021unified} or graph based approaches \cite{xu-etal-2020-user,lei2020interactive,ren2021learning,xu2021adapting}.

\subsubsection{Strategies Learning in Multi-turn CRS}
Some works in this category focus on balancing the trade-off between exploration (i.e., asking questions) and exploitation (i.e., making recommendations), especially for cold-start users. They study the trade-off strategies to achieve engaging and successful recommendations. Some of them \cite{li2010contextual,li2016collaborative,christakopoulou2016towards,li2020seamlessly} leverage bandit online recommendation methods and focus on cold-start scenarios, while others work on strategically asking clarification questions with fewer turns~\cite{lei2020estimation,lei2020interactive,sun2018conversational}. 

\subsubsection{Open-ended CRS} 
An open-ended CRS aims to make recommendation in a more natural and casual way compared with the task-oriented CRS. Many datasets have been collected or built to push forward the research of CRS, including ReDial \cite{li2018towards}, TG-ReDial (Chinese) \cite{zhou2020towards}, GoRecDial \cite{kang2019recommendation}, DuRecDial (Chinese) \cite{liu2020towards}, INSPIRED \cite{hayati2020inspired} and OpenDialKG \cite{moon2019opendialkg} datasets.  
Most of them consist of around 10,000 conversations that are focused on recommendation and chit-chat on different domains. For example, ReDial is about movie recommendation, while OpenDialKG is concerned with several domains, including movie, book, sports and music. The follow-up studies on the ReDial dataset generally divide the CRS into recommendation and generation modules. For the recommendation module, the previous works tend to apply more and more external knowledge to improve the recommendation performance, e.g., entity-level knowledge graph \cite{chen-etal-2019-towards}, word-level knowledge graph \cite{zhou2020improving} and item reviews \cite{lu2021revcore}. However, it's challenging to manage so much external knowledge via an end-to-end model. What's more, some (like item reviews) might need much effort to collect and annotate, and are not generally applicable for all kinds of domains 
(e.g., some items might lack reviews). 
For generation module, most of the previous works adopt encoder-decoder framework and train the generation model from scratch. However, it's difficult to learn diverse and valuable patterns from relatively small datasets. Our work further explores approaches and improves quality of CRS for this category.

\subsection{Recommender System}
Traditional recommender system is different from the conversational recommender system, as it generally does not focus on dialogue interactions with users. It makes more efforts on the information filtering techniques to handle information overflow problems in recent years. 

The problems in the traditional recommender system can be divided into two categories \cite{batmaz2019review}: recommendation with explicit feedback (i.e., the ratings given to products) and recommendation with implicit feedback (purchase history or click-through history) \cite{hu2008collaborative}. The former problem can be evaluated directly by the differences between ground truths and predictions while the evaluation of the latter problem is usually formulated as Top-N ranking problems. 

The methods proposed in the conventional recommender system can be divided into three categories: collaborative filtering based recommendation \cite{schafer2007collaborative}, content-based recommendation \cite{van2000using} and the hybrid of them \cite{tran2000hybrid}. Collaborative filtering has been the most popular recommendation technique in recent years. It hypothesizes that people who have had similar interests before would also prefer similarly in the future \cite{sarwar2001item}. Many methods are proposed to identify the similarity of users or items, including memory-based methods \cite{breese2013empirical}, clustering methods \cite{ungar1998clustering}, Bayesian models \cite{su2006collaborative} and matrix factorization methods \cite{koren2009matrix}. However, collaborative filtering methods may perform poorly when handling new items that lack history information. As for content-based recommendation methods, they usually first compare the similarity between the descriptive characteristics of items and the user profiles, and then recommend items that are similar to what users like before. Therefore, the content-based methods are usually efficient in recommending new items. But they are limited in terms of diversity and novelty \cite{lops2011content}. The hybrid of collaborative filtering and content-based methods take advantages of both of them \cite{tran2000hybrid}.

\subsection{Pre-trained Language Model}
The pre-trained language models \cite{devlin-etal-2019-bert,liu2019roberta,lewis-etal-2020-bart,brown2020language} become more and more popular in recent years. The idea behind them is first pre-training the models on large-scale unlabeled corpus (which can be easily crawled on the Internet) and then apply them to many downstream tasks (e.g., QA system, dialogue system, summarization and machine translation). The architecture of these pre-trained language models has been developed from shallow to deep (the parameter scale is advancing from million-level \cite{devlin-etal-2019-bert,brown2020language} to trillion-level \cite{fedus2021switch}) along with the improvement of computational capability and the enhancement of training mechanisms. Pre-training tasks such as nearby word prediction  \cite{mikolov2013distributed}, next word prediction  \cite{bengio2003neural,wang2015unsupervised}, masked language model \cite{devlin-etal-2019-bert} are widely explored in language modeling. These tasks do not need any annotated labels, so they can be trained on a huge amount of unlabeled data from the Internet, e.g., Wiki and Reddit.

We introduce how these pre-trained language models are applied to the downstream tasks. The development of utilizing pre-trained language models can be divided into two stages, embedding-based approaches and finetuning-based approaches. In embedding-based approaches, the word-, sentence- or paragraph-level embeddings from the pre-trained language models is utilized as features in the downstream tasks. As for finetuning-based approaches, they benefit the knowledge transferring between the pre-trained language models and the downstream tasks. A common finetuning procedure would fix (or apply smaller learning rate to) the original parameters of the pre-trained language model and add some finetunable adaptation modules for the downstream tasks. Besides, prompt tuning is getting more attention in all of finetuning methods. Discrete \cite{petroni-etal-2019-language,gao2020making} or continuous \cite{liu2021gpt} prompts based finetuning are proposed to help bridge the gap between the pre-training and finetuning and reduce the computational cost. 
Different from the domain-specific knowledge (e.g., review of items) introduced by the previous work \cite{lu2021revcore,zhou2022c2}, the information learned by pre-trained language models is general and not task-specific. The finetuning methods based on pre-trained language model can be well generalized to other domains while the training scheme based on domain-specific knowledge limits the model generalizability.

\section{Methodology}\label{sec:method}
\begin{figure*}[th]
\centering
\includegraphics[width=1.0\textwidth]{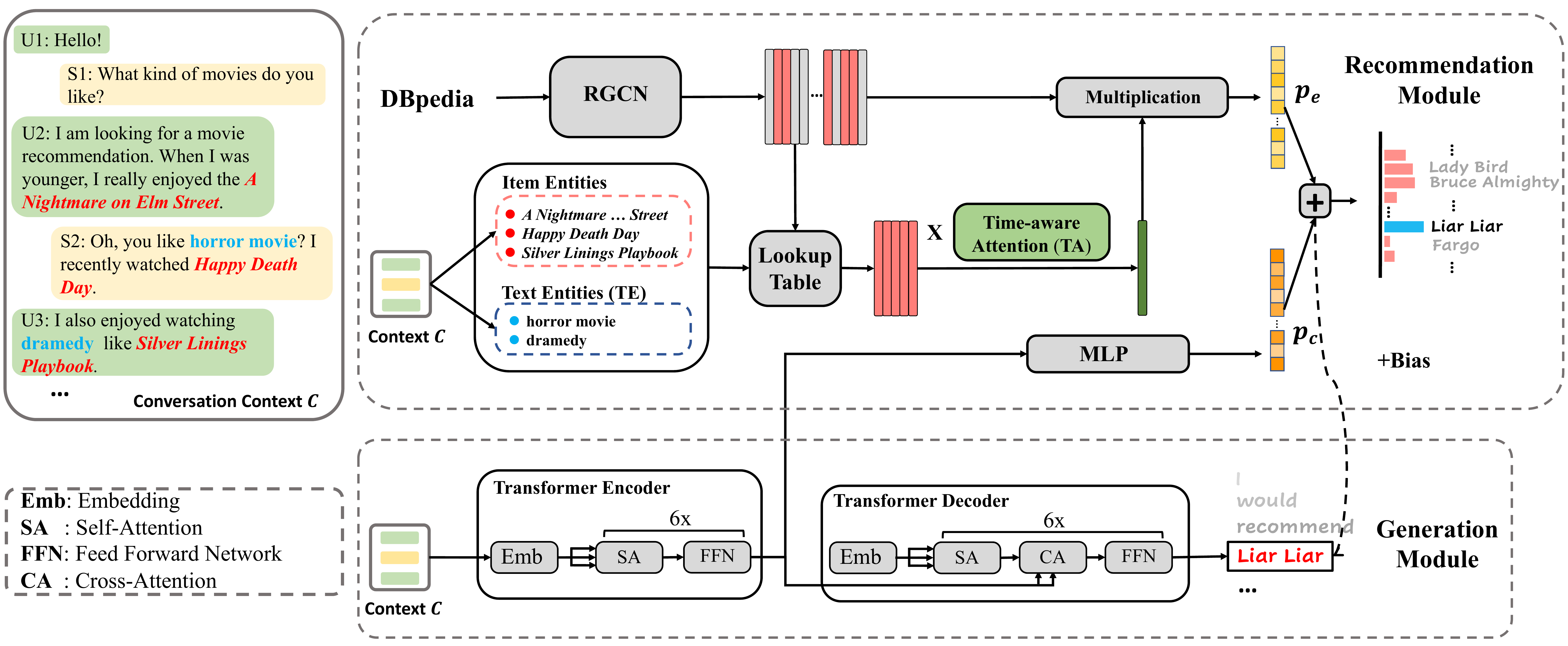}
\vskip -0.5em
\caption{\label{fig:sketch} Our framework for conversational recommendation. It consists of two modules, a Recommendation Module to rank the candidate items and a Generation Module to generate fluent responses to support interaction with users. The input of our framework can be two parts: a knowledge graph and context $C$. The outputs are responses with recommended items.}
% \vskip -1.5em
\end{figure*}
In this section, we first formulate the task of CRS in Section \ref{sec:model:problem}, followed by the description of our generation module in Section \ref{sec:model:generation}, which is finetuned with a pre-trained sequence-to-sequence framework BART~\cite{lewis-etal-2020-bart}. Then we introduce how we do recommendation based on conversation context and entity-level information in Section \ref{sec:model:recommendation}. Finally, we describe how we integrate the above two modules and produce the final responses in Section \ref{sec:model:integration}.

\subsection{Problem Formulation}
\label{sec:model:problem}
A classical architecture of conversational recommender system generally consists of two modules, named \emph{generation} module and \emph{recommendation} module. 
It takes the history context of a conversation $C = (t_1, t_2, \ldots, t_m)$ as input, where $t_i$ ($1 \le i \le m$) is an utterance from the seeker (i.e. the user) or the CRS itself and $m$ is the number of context utterances. Each utterance $t_i$ contains word tokens $t_i = (w_{i,1}, w_{i,2}, \ldots, w_{i,n_i})$, where $n_i$ is the number of tokens in utterance $t_i$. 
The CRS uses its recommendation module to recommend items $\mathcal{I}_i$ from a candidate item set $\mathcal{I}$ based on the context $C$ and some related external knowledge (e.g., knowledge graphs or item reviews). 
Then it embeds them into a response $R = (y_1, y_2, \ldots, y_{n})$, a sequence of $n$ tokens generated by the generation module based on the context $C$ to produce natural responses with recommendations.

\subsection{BART-based Response Generation}
\label{sec:model:generation}

Our response generation module follows a general Transformer~\cite{DBLP:conf/nips/VaswaniSPUJGKP17} sequence-to-sequence framework. We first briefly introduce the Transformer framework followed by the BART pre-trained language model. Then we explain how we finetune the BART model to enable generating responses with recommended items.

\textbf{Transformer.} 
A Transformer layer typically consists of two kinds of sub-layers: multi-head attention layers and fully connected feed-forward networks. In a multi-head attention layer, for each head $j$, the output vector is computed with the following equation:
\begin{equation}
{\mathop{\mathrm{Attn}}}_{j} (\bm{Q}, \bm{K}, \bm{V}) = \softmax (\frac{\bm{Q}_j\bm{K}_j^T}{\sqrt{d_k}}) \bm{V}_j
\end{equation}
where $\bm{Q}, \bm{K}, \bm{V}$ are the matrices of queries, keys and values, $d_k$ is the dimension of keys, and $\bm{Q}_j = \bm{Q}\bm{W}_j^Q$, $\bm{K}_j = \bm{K}\bm{W}_j^K$, $\bm{V}_j = \bm{V}\bm{W}_j^V$ are project matrices for head $j$. The output of each head is concatenated and then mapped to the final output field. For the fully connected feed-forward network (${\mathop{\mathrm{FFN}}}$), it consists of two linear transformations with an ReLU activation in between:
\begin{equation}
\mathop{\mathrm{FFN}} (\bm{X}) = \relu (\bm{X}\bm{W}_1 + \bm{b}_1)\bm{W}_2 + \bm{b}_2
\end{equation}

A Transformer encoder layer is composed of a multi-head self-attention layer (takes the same input as $\bm{Q}, \bm{K}, \bm{V}$) and a feed-forward network, while a Transformer decoder layer additionally attends to the encoder output (as $\bm{K}$ and $\bm{V}$) for cross attention (see the bottom part of \Cref{fig:sketch}).

In summary, the whole process for a general Transformer sequence-to-sequence framework is as follows:
\begin{align}
\label{eq:enc}\bm{H}^{C} &= \enc(x)  \\
y_k &= \dec(y_{<k},\bm{H}^{C}) \\
\mathcal{L}_{gen} &= \sum\nolimits_{k=1}^{n} - \log (p(y_k|y_{<k},\bm{H}^{C})) \label{eq:transformer-gen}
\end{align}
\noindent where $x$ is the input token sequence, and
$y_{<k}$ represents the target tokens before $y_k$. The $\enc$ includes a word embedding layer and several Transformer encoder layers while the $\dec$ is similar but additionally includes an output projection layer. The model is learned to minimize the negative log-likelihood of the target sequence as in~\Cref{eq:transformer-gen}.

\textbf{BART.}
BART~\cite{lewis-etal-2020-bart} is a pre-trained language model based on Transformer trained with several denoising objectives on large-scale books and Wikipedia data, i.e., recovering the noisy input processed by token masking, token deletion, text infilling, sentence permutation and document rotation. These denoising objectives enable the model to learn contextual representations with unlabeled data, which can be crawled easily from the Internet, and then used for multiple downstream tasks. BART has been shown to be effective in many generation tasks, including abstractive QA, summarization, machine translation at the sentence and document level, and persona-based response generation. 

\textbf{Finetune BART for CRS.}
As most of the available CRS datasets are relatively small and contain only around 10K conversations~\cite{li2018towards,moon2019opendialkg}, it is difficult to learn complex semantic and discourse level dependencies only based on the training corpus. To relieve the burden and enhance context modeling, we choose to finetune a pre-trained BART~\cite{lewis-etal-2020-bart} model for our response generation module. Though the introduction of pre-trained BART model can be regarded as a way of including more external knowledge to the CRS, there are two major differences between the domain-specific external knowledge (i.e., review of items) and the general external knowledge (i.e., pre-trained language model): (i) Difficulties in applying the knowledge. The collection of domain-specific knowledge is time-consuming and how to design efficient networks to include the domain-specific knowledge with other knowledge becomes a new challenge, while the finetuning scheme based on the pre-trained language model is easier and can follow some regular solutions. (ii) Generalizability. The pre-trained language models can be employed in most of the tasks while the domain-specific knowledge may only exist in a few tasks or be hard to collect for some tasks. 

To enable the BART model to generate the item-aware responses, we extend its original vocabulary $\mathcal{V}$ with the item set $\mathcal{I}$ to be $\mathcal{V}^{\prime} = \mathcal{V} \cup \mathcal{I}$ (in this case, we do not tokenize the item names in an utterance and recognize them as single tokens), and then use a CRS training corpus to finetune the model. 
Specifically, we concatenate the utterances $t_i$ in context $C$ with an appended $\langle EOT\rangle$ token in their chronological order as the input (i.e., $x = [t_1;\langle EOT\rangle;t_2;..;\langle EOT\rangle;t_m]$), and maximize the probability of the ground truth response $R$ (i.e., minimize the objective in ~\Cref{eq:transformer-gen}).

Finally, an integration operation is added to the generation output to make the generation aware of the recommendation. We will discuss it later in Section \ref{sec:model:integration}.

\subsection{Context-Time-Aware Recommendation}
\label{sec:model:recommendation}
To fully understand user preferences over items from a given context $C$, we propose to extract two kinds of information for the recommendation. The first is entity-level information, where we extract the mentioned entities (including the items in the item set $\mathcal{I}$) from $C$ and apply them to an external relational knowledge graph to perform entity linking~\cite{daiber2013improving}. Then the learned entity representations are summarized by a proposed time-aware attention to produce entity-level user representation.
The second is contextual information produced by BART model, which is expected to capture information from the perspectives of semantic and conversational discourse. We describe the details in the following.

\subsubsection{\textbf{Entity-level Representation}} 
\label{sec:model:rec:entity}
Entity-level representation is captured and represented by extracted items as well as their related entities (e.g., if the items are the movies, the related entities can be the authors, directors, etc.) from the observed context $C$. Because of the data scarcity, it is not promising to summarize the extracted entities merely based on the training corpus~\cite{chen-etal-2019-towards}. 
We follow previous work and employ a relational knowledge graph (e.g. DBpedia) to enhance entity modeling. 

\textbf{Entity Linking.} 
Specifically, we denote a triplet in the relational knowledge graph with $\langle e_1, r, e_2\rangle$, where $e_1, e_2 \in \mathcal{E}$ are entities from the entity set $\mathcal{E}$ and $r$ is a kind of entity relation from the relation set $\mathcal{R}$. For the ReDial dataset, we utilize the linked knowledge graph, a subset of DBpedia, which is released by \cite{chen-etal-2019-towards}. For the OpenDialKG dataset, we also extract a subset of DBpedia by matching each item in the item set to entities in DBpedia and additionally linking them on dialogue contents by following \cite{daiber2013improving}. 

We use an R-GCN~\cite{schlichtkrull2018modeling} framework to encode relation-aware entity representations. The intuition behind this is that  
the features of an entity can be indicated and summarized by its neighboring nodes in the knowledge graph.
An R-GCN computes the embeddings of the entities with multi-edge encoding with several iteration layers. Formally, the representation of an entity $e$ at the $(l+1)$-th layer is:
% calculated as follows:
\begin{equation}
\setlength\belowdisplayskip{1pt}
\bm{h}_e^{(l+1)} = \relu (\sum_{r \in \mathcal{R}^{\prime}} \sum_{e^{\prime}\in\mathcal{E}_e^r} \frac{1}{Z_{e,r}}\bm{W}_r^{(l)}\bm{h}_{e^{\prime}}^{(l)})
\end{equation}
\noindent where $\bm{h}_e^{(l)}$ represents the representation of entity $e$ at the $l$-th layer, and $\mathcal{E}_e^r$ denotes the set of neighboring nodes of $e$ under the relation $r$. $\mathcal{R}^{\prime} = \mathcal{R} \cup \{r_{self}\}$ contains all the relations including self loop.
$\bm{W}_r^{(l)}$ is a learnable relation-specific transformation matrix and $Z_{e,r}$ is a normalization factor.
Therefore, for each node in the knowledge graph, it receives and aggregates the messages from its neighboring nodes after relation-specific transformation. Then it combines the information with the hidden representations to form its updated representation at the next layer. Finally, structural and relational information is encoded into the entity representation $\bm{h}_e^{L}$ for each $e \in \mathcal{E}$ at the last layer $L$. For simplicity, we represent the representations in the final layer $L$ as $\bm{h}_e$ by omitting the superscript ``$L$''.

\textbf{User Preference Summarization.}
Given a context $C$, we presume we can summarize the preference of the seeker with the appeared entities.
We extract the appeared entities (as user preference) $\mathcal{T}_u = \{e_1, e_2, ..., e_{|\mathcal{T}_u|}\}$ from two perspectives: item entities (i.e., entities that appear in item set $\mathcal{I}$) and other relevant contextual entities (mentioned in utterances but not an item in $\mathcal{I}$, e.g., an actor of a movie item. 
We denote them as \textit{text entities}). For items that are not covered by $\mathcal{E}$, we ignore them by following the previous work \cite{chen-etal-2019-towards,zhou2020improving,lu2021revcore}. The entities $e_i \in \mathcal{E}$ are sorted in the order of appearance. After looking up the entities in $\mathcal{T}_u$ from $\bm{H} = \{\bm{h}_e\}_{e=1}^{|\mathcal{E}|}$, we get $(\bm{h}_1, \bm{h}_2, ..., \bm{h}_{|\mathcal{T}_u|})$.

To summarize the entity-level user representation, previous work mainly depends on a self-attention mechanism~\cite{zhou2020improving,lu2021revcore}, where a learnable matrix is used to learn and derive each entity's importance. Such a naive mechanism might be sub-optimal for the following two reasons: \textbf{(i)} no supervised signals are used to guide the model to learn knowledge about entity importance which makes it difficult to summarize accurately, especially when the training corpus is limited; \textbf{(ii)} it ignores the order of appearance which might affect the conversation trend (e.g. the appeared entities $A, B, C$ and $C, B, A$ probably lead to different next items in a conversation, but we will get the same summarized representation with such a mechanism). Therefore, we propose \textbf{Time-aware Attention} to address the limitations, where the entity-level user representation $\bm{h}^E$ is calculated as follows:

\begin{equation}
\label{eq:time}
\bm{h}^E = \sum_{i=1}^{|\mathcal{T}_u|} \frac{\lambda^{i-1}}{\sum_{i=1}^{|\mathcal{T}_u|} \lambda^{i-1}} \bm{h}_i
\end{equation}
\noindent where $\lambda$ is a hyper-parameter to control the recency effect. The value of $\lambda$ is usually larger than $1$, which means that the recently appeared items will contribute more to the next item prediction. The intuition behind the time-aware attention can be twofold: \textbf{(i)} The currently available conversational recommendation datasets are quite small and the previous proposed self-attention based item representation could not show promising performance (see discussion in Section \ref{sec:exp_res:rec:redial}). Our simple but effective modeling mechanism can handle multi-domain datasets regardless of their sizes. \textbf{(ii)} One of the goals of conversational recommender system is to guide the users to better express their preferences explicitly through the ongoing conversations. Therefore, it's naturally to put more weights on the most recently appeared items. We will discuss the effect of the value of $\lambda$ in Section~\ref{sssec:para_effects1}.

Finally, the time-aware entity-level recommendation probability can be computed as follows: 
\begin{equation}
\bm{p}_{e} = \softmax (\mask(\bm{h^E} \bm{H}^\top))
\end{equation}
\noindent where $\mask$ is an operation that sets all non-item entities to $-\infty$ (to make the recommendation focus on the candidate items in item set $\mathcal{I}$), and $\bm{p}_{e} \in \mathbb{R}^{|\mathcal{E}|}$. $\bm{p}_{e}$ will be used for recommendation together with contextual-level representation. We introduce the latter one in the next subsection.

\subsubsection{\textbf{Contextual-level Representation}}
\label{sec:model:rec:context}
Though the time-aware entity-level representation can model the user preference from the perspective of the mentioned items in context $C$, it may lead to misunderstanding of user preferences since it ignores the content from the context. 
For example, if a user says ``I do not like A!''. We cannot capture such a negative opinion towards ``A'' through entity-level representation alone. 

To partially address the problem and to incorporate more semantic- and discourse-level context for recommendation, we further use the context representation $\bm{H}^{C}$ computed in \Cref{eq:enc} from generation module to yield semantic-aware prediction. Specifically, we average the context representations over the sequence as contextual-level representation for $C$ and put it through an MLP layer to give the prediction:
\begin{equation}
\bm{p}_{c} = \softmax (\mlp(\frac{1}{|C|}\sum_{j=1}^{|C|} \bm{h}_{j}^{C}))
\end{equation}
\noindent where $\bm{h}_{j}^{C}$ denotes the representation for the $j$-th token in the context representation $\bm{H}^{C}$, $|C|$ is the total length of context, and $\bm{p}_{c} \in \mathbb{R}^{|\mathcal{E}|}$. 
$\bm{p}_{c}$ reflects the effects from the content of context for recommendation, which can be a complement for entity-level representation. We next describe how we combine these two kinds of representations.

\subsubsection{\textbf{Combination of the Entity- and Contextual-Level Recommendation}} 
Different from the previous work, our work considers both entity-level representation and contextual-level representation for recommendation. 
The final recommendation based on the above two components is defined as:
% defined as:
% \vskip -1em
\begin{equation}
\label{eq:rec-joint}
\bm{p}_{rec} = \mu \cdot \bm{p}_e + (1 - \mu) \cdot \bm{p}_c
\end{equation}
% \vskip -0.5em
\noindent where $\mu$ ($0 \leq \mu \leq 1$) is a hyper-parameter to balance between the two kinds of recommendations. Therefore, the learning objective for the recommendation module is:
% can be summarized as:
% \vskip -1em
\begin{equation} 
    \mathcal{L}_{rec} = - \sum_{i=1}^{M} \log p_e(r_{i}) + \log p_c(r_{i})
\end{equation}
% \vskip -0.5em
\noindent where $M$ is the number of items extracted from the training corpus that need to be recommended, while $r_{i} \in \mathcal{I}$ is the target item in the $i$-th recommendation. $p_e(r_{i})$ and $p_c(r_{i})$ are the corresponding prediction probabilities of the target item from the entity-level and contextual-level recommendation components, respectively.

\subsection{Module Integration}
\label{sec:model:integration} 

So far, the responses generated by the generation module (introduced in Section \ref{sec:model:generation}) are not aware of the recommendation (introduced in Section \ref{sec:model:recommendation}) results. 
We introduce an integration mechanism to incorporate the recommendation module's results and guide the generation module to generate responses that are consistent with the user’s preference. 
Some of the previous works~\cite{zhou2020improving,lu2021revcore} use copy mechanism~\cite{gu-etal-2016-incorporating,gulcehre-etal-2016-pointing} to incorporate recommendation results into the generated responses. However, as it needs additional networks to predict when to switch between generation mode and copy mode, it may sometimes perform wrongly due to insufficient data to learn the pattern.

Inspired by \cite{chen-etal-2019-towards}, we use a simpler but effective integration mechanism. We directly add a \emph{vocabulary bias} to the top decoder predictions. Different from their work, our vocabulary bias directly comes from the recommendation probabilities $\bm{p}_{rec}$ in~\Cref{eq:rec-joint}:
% \vskip -1em
\begin{equation} 
    \bm{b}_u = [\bm{0};\mathcal{G}(\bm{p}_{rec})]
\end{equation}
% \vskip -0.5em
\noindent where $\bm{0}$ is a $\mathcal{V}$-dimensional zero vector, $\mathcal{G}(\cdot)$ is an index selection operation to select items from entities ($\mathcal{G}:\mathbb{R}^{|\mathcal{E}|}\rightarrow\mathbb{R}^{|\mathcal{I}|}$), and $[;]$ means concatenation. This makes the bias has the same dimension as our generation output (i.e. $|\mathcal{V}^{\prime}| = |\mathcal{V}| + |\mathcal{I}|$). 

We dynamically add the bias $\bm{b}_u$ during generation based on the top predictions in each time stamp $t$. Only when the top predictions of that time stamp include items in item set $\mathcal{I}$ will we add the bias to them. The effect of this operation is that for a generation token $y_t$ that is probably an item in $\mathcal{I}$, we add its recommendation probability $p_{rec} (y_t)$
to the original generation probability $p(y_t)$. In this way, the results of the two modules are integrated so that our system can generate the recommendation-aware responses.

The total objective for our model is: 
% defined as follows:
% \vskip -1em
\begin{equation} 
\label{eq:final-loss}
    \mathcal{L} = \mathcal{L}_{gen} + \gamma \mathcal{L}_{rec}
\end{equation}
% \vskip -0.5em
\noindent where $\gamma$ is a hyper-parameter that balances the two objectives. We jointly train the two objectives so that the two modules can benefit each other.

\section{Experimental Setup}\label{sec:exp_setup}
\begin{table}[t]\setlength{\tabcolsep}{1.5mm}
\caption{\label{tab:statistic} Statistics of ReDial and OpenDialKG datasets.
}
% \vskip -1em
\begin{center}
\resizebox{\linewidth}{!}{
\begin{tabular}{l|cc}
\toprule[1.0pt]
& \textbf{ReDial}  &\textbf{OpenDialKG} \\
\midrule[0.5pt]
Number of conversations &10,006 &13,802 \\
Number of utterances &182,150 &91,209 \\
Avg utterance number per conv & 18.2 & 6.6 \\
Knowledge Graph &DBpedia &DBpedia \\
Domains & Movie & Movie, Book, Sports, Music  \\
% \midrule[0.5pt]
% Number

\bottomrule[1.0pt]

\end{tabular}}
\end{center}
% \vskip -1em

% \vskip -1.5em
\end{table}
In this section, we first show the basic statistics of the two used datasets and the evaluation metrics for recommendation and generation modules in Section \ref{ssec:datasets_eval}. Then we describe the baseline models together with some variants to our main model in Section \ref{ssec:baseline_variants}. Finally, we list the parameter settings of our model in Section \ref{ssec:parameter_setting}.

\subsection{Datasets and Evaluation}\label{ssec:datasets_eval}
\textbf{Datasets.} To empirically evaluate the proposed approach, we conduct experiments on two popular CRS datasets, namely, ReDial~\cite{li2018towards} and OpenDialKG~\cite{moon2019opendialkg}. 
ReDial is centered around movie recommendation and is constructed on Amazon Mechanical Turk (AMT) platform with pair crowd-workers (i.e., Seeker and Recommender). OpenDialKG consists of conversations that are mainly in four domains: movie, book, sports and music. The construction process of OpenDialKG is described as follows, which can reveal why the recommendation on OpenDialKG is much easier than on ReDial: The first agent initiates a conversation by giving a seed entity. The second agent is given with a list of facts that are relevant to the seed entity, and asked to choose the most relevant facts and use them to frame a free-form conversational response. 
\Cref{tab:statistic} presents basic statistics about the two datasets. As can be seen, the both datasets have around 10K conversations. 
We can notice that the average conversation length of ReDial dataset is larger than that of OpenDialKG. For the experiments, we split the two datasets into training, validation and test at ratios of 80\%:10\%:10\% and 75\%:15\%:15\% by following \cite{li2018towards} and \cite{moon2019opendialkg}, respectively.

\textbf{Evaluation Metrics.} 
We evaluate the performance of recommendation and generation separately. For recommendation, we adopt Recall@K scores where K = 1, 10, 50 for ReDial by following \cite{chen-etal-2019-towards}, and K = 1, 3, 5, 10, 25 for OpenDialKG by following \cite{moon2019opendialkg}.  Recall@K indicates whether the predicted top-K items contain the ground truth recommendation items. For generation, apart from Dist-n (n=2, 3, 4) and perplexity (PPL) scores reported in previous work, we also report the case-insensitive BLEU-n (n=2, 4) scores\footnote{We use NLTK package (\url{https://www.nltk.org}) to calculate the BLEU scores.}. For a fair comparison, we calculate the PPL scores via a widely used off-the-shelf package KenLM\footnote{\url{https://kheafield.com/code/kenlm/}}, as the PPL scores may be very different when using different vocabulary.

\subsection{Baselines and Variants of Our Model}\label{ssec:baseline_variants}
For the ReDial dataset, we compare our model with six competitive baselines proposed by the previous work: \\ 
\textbf{$\bullet$ ReDial} \cite{li2018towards}: The model consists of a dialog generation module which is based on HRED \cite{sordoni2015hierarchical}, a recommendation module that is based on an auto-encoder \cite{he2017distributed} and a sentiment analysis module. Switching Mechanism is adopted to combine the recommendation and generation results. \\
\textbf{$\bullet$ KBRD} \cite{chen-etal-2019-towards}: It utilizes DBpedia knowledge graph to enhance the semantic information of items (or entities) to do the recommendation and adopts a transformer based generation module, where knowledge graph information serves as word bias to assist the generation. \\
\textbf{$\bullet$ CRWalker} \cite{ma2020bridging}: The method walks on the knowledge graph to form a reasoning tree at each turn for recommendation, and then maps to dialog acts to guide response generation. The learned recommendation-oriented dialog policy on the knowledge graph enhances the mutual benefit between recommendation and conversation. \\
\textbf{$\bullet$ KGSF} \cite{zhou2020improving}: It uses mutual information maximization (MIM) \cite{viola1997alignment} to align the semantic spaces of word- and entity-level KGs for the recommendation module. Its generation module includes a transformer encoder followed by a fused knowledge-enhanced decoder. \\
\textbf{$\bullet$ RevCore} \cite{lu2021revcore}: It performs review-enriched and entity-based recommendation and uses a review-attentive encoder-decoder for generation. We use the re-implementation results in \cite{zhou2022c2} for a fair comparison. \\
\textbf{$\bullet$ $C^2$-CRS} \cite{zhou2022c2}: This method extracts and represents the associated multi-grained semantic units from different data signals, and then align the corresponding semantic units from different data signals in a coarse-to-fine way. 

For the OpenDialKG dataset, we compare the models that are compared in \cite{moon2019opendialkg} and introduce them as follows.\\
\textbf{$\bullet$ seq2seq} \cite{sutskever2014sequence}: It applies a seq2seq approach for entity path generation, given all of the dialog contexts.\\
\textbf{$\bullet$ Tri-LSTM} \cite{young1709augmenting}: It encodes each utterance and all of its related facts within 1-hop from a KG to retrieve a response from a small (N=10) pre-defined sentence bank. \\
\textbf{$\bullet$ Ext-ED} \cite{parthasarathi2018extending}: It generates response entity token  at its final softmax layer, and does not utilize structural information from knowledge graph. \\
\textbf{$\bullet$ DialKG Walker} \cite{moon2019opendialkg}: It has an attention-based graph decoder that penalizes decoding of unnatural paths which prunes candidate entities and paths from a large search space. \\

We also test the performance of the following variants to our model as an ablation study. \\
\textbf{$\bullet$ \textsc{BART (ContextM)}}: It only uses contextual-level representation (described in Section \ref{sec:model:rec:context}) to do the recommendation. No entity-level representations and extra knowledge graph based knowledge are employed. \\
\textbf{$\bullet$ \textsc{EntityM-SelfA}}: \textsc{EntityM-SelfA} refers to entity modeling with self-attention used in previous work. It only uses the entity-level representations to do the recommendation.\\
\textbf{$\bullet$ \textsc{EntityM-TimeA}}: \textsc{EntityM-TimeA} refers to entity modeling with our designed time-aware modeling. It also only uses entity-level representations. \\ 
\textbf{$\bullet$ \textsc{EntityM-TimeA-ContextM}}: Our full model that adopts both entity-level and contextual-level representation.
% to do the recommendation. 

Besides, to examine the effectiveness of adding text entities to input (defined in Section \ref{sec:model:rec:entity}) for recommendation, we also evaluate our model variants with and without text entities (i.e., TextEntity in \cref{tab:results_rec_redial}) as input.

\subsection{Parameter Setting} \label{ssec:parameter_setting}
% \paragraph{\textbf{Parameter Setting.}} 
For the RGCN-based recommendation module, we set both the entity embedding size and the hidden representation dimension to 128. The layer number for R-GCN is 1 and the normalization factor $Z_{e,r}$ is set to 1 following \cite{chen-etal-2019-towards}. For the BART-based generation module, we finetune on the pre-trained BART-Base model, which consists of 6 layers of encoder and decoder, respectively. The hidden dimension, feed-forward network size and attention head number are 768, 3072 and 12 for both encoder and decoder layers. The total parameter number of our model is 269M. The time recency effect $\lambda$ in \Cref{eq:time}, the balance factor between entity- and contextual-level representations in \Cref{eq:rec-joint} and the tradeoff between two training objectives in \Cref{eq:final-loss} are set to 1.5, 0.5 and 1.0, respectively. We adopt diverse beam search~\cite{vijayakumar2016diverse} mechanism in generation with a beam size of 4 and the diverse beam group number is set to 2. All the hyper-parameters are determined by grid-search based on the validation performance. 

We implement our models based on FAIRSEQ framework\footnote{\url{https://github.com/pytorch/fairseq}} \cite{ott2019fairseq}, and train on an NVIDIA 3090 GPU. 
During training, we set the max tokens of each batch to 4096 with an update frequency of 4. We adopt Adam optimizer with a 5e-3 learning rate (and 5e-5 learning rate for BART-related modules as they are pre-trained parameters) 
% since it will be pre-trained on the conversations first) 
and 1000 warm-up updates followed by a polynomial decay scheduler.
The training time of one epoch is around 22 minutes. The model needs around 5 epochs to achieve the best performance on the validation set.

\begin{table*}[t]
\caption{\label{tab:results_rec_redial} \textbf{Recommendation} results (in \%) on ReDial dataset. Our full model achieves the best performance with less external domain-specific knowledge as input, compared to the previous work.
}
\begin{center}
\begin{tabular}{l|ccccc|ccc}
\toprule[1.0pt]
\multirow{2}{*}{\textbf{Models}} & \multicolumn{5}{c|}{\textbf{Input}} & \multirow{2}{*}{\textbf{Rec@1}}  & \multirow{2}{*}{\textbf{Rec@10}} & \multirow{2}{*}{\textbf{Rec@50}} \\
\cmidrule(lr){2-6}
& Context & TextEntity &DBPedia &ConceptNet &Review & &&\\
\midrule[0.5pt]
\underline{Baselines} & & & & & & & &  \\
\textsc{ReDial} &\Checkmark & & &  && 2.4  &14.0 & 32.0  \\
\textsc{KBRD} &\Checkmark &\Checkmark &\Checkmark &  & & 3.1 &15.0 &33.6 \\
\textsc{CRWalker} &\Checkmark &\Checkmark &\Checkmark & & & 3.1 & 15.5 & 36.5 \\
\textsc{KGSF} &\Checkmark &\Checkmark &\Checkmark &\Checkmark  & & 3.9 & 18.3 & 37.8 \\
\textsc{RevCore} &\Checkmark  &\Checkmark&\Checkmark  & &\Checkmark & 4.6 & 22.0 & 39.6  \\
\textsc{$C^2-$CRS} &\Checkmark &\Checkmark &\Checkmark & &\Checkmark &5.3 &23.3 &40.7 \\

\midrule[0.5pt]
\underline{Our Models} & & & &  &  & & & \\
\textsc{BART(ContextM)} &\Checkmark & & & &  & 3.0 &16.4  &35.0 \\
\textsc{EntityM-TimeA} &\Checkmark & &\Checkmark& & & 4.6 &18.3 &34.1 \\
\textsc{EntityM-TimeA-ContextM} &\Checkmark & &\Checkmark& &  & 5.7&22.6&40.4 \\
\textsc{EntityM-SelfA}&\Checkmark &\Checkmark &\Checkmark& &  & 3.3 & 16.3 & 32.6 \\
\textsc{EntityM-TimeA} &\Checkmark &\Checkmark &\Checkmark& &  & 5.2 &18.2 &34.6 \\
\textsc{EntityM-TimeA-ContextM} &\Checkmark &\Checkmark &\Checkmark& &  & \textbf{5.9}  &\textbf{24.0} &\textbf{41.3} \\
\bottomrule[1.0pt]
\end{tabular}
% }
\end{center}
% \vskip -1em
\vskip -1em
\end{table*}
\section{Experimental Results}\label{sec:exp_res}
In this section, we first report the main comparison results on recommendation and generation in Section \ref{sec:exp_res:rec} and Section \ref{sec:exp_res:gen}, respectively. Then we analyze the effectiveness of our proposed modules in Section \ref{ssec:exp:effectiveness} followed by a case study in Section \ref{sec:exp_res:case}. Finally, we give more analysis in Section \ref{ssec:further_analysis}.

\subsection{Recommendation Result Comparisons}\label{sec:exp_res:rec}
In this subsection, we present the main comparison results on ReDial dataset in Section \ref{sec:exp_res:rec:redial}. Then, to further verify the effectiveness of our model in multi-domain dataset, we conduct experiments on OpenDialKG dataset and report the comparison results in Section \ref{sec:exp_res:rec:opendial}.

\subsubsection{\textbf{Results on ReDial}}\label{sec:exp_res:rec:redial}
\Cref{tab:results_rec_redial} shows the recommendation results of the baselines and the variants of our model on ReDial Dataset. 
We can draw the following observations from the results:

$\bullet$~\textit{Previous work tends to add more external knowledge, which can improve the recommendation performance, but ignores fully extracting the internal information.} We can see that among the baselines, \textsc{KBRD} and \textsc{CRWalker} add the entity-level knowledge (EntityKG, i.e., DBPedia and the text entities extracted from it) as input. \textsc{KGSF} further introduces word-level knowledge (WordKG, i.e., ConceptNet),
while \textsc{RevCore} and \textsc{$C^2-$CRS} further incorporate item reviews (Review). All the added external knowledge introduces considerable improvement, demonstrating the effectiveness of the external knowledge. However, we can also notice that our \textsc{EntityM-TimeA} (Entity Modeling with Time-aware Attention) model only with the input of ``Context'' and ``DBPedia'' can also get a $4.6 \%$ Recall@1 score which is comparable with the results of \textsc{RevCove}, who utilizes two more external knowledge (Text Entity and Review). This shows that the previous designed methods cannot fully extract the internal knowledge from context, but our proposed model can capture more useful information for recommendation.

$\bullet$~\textit{Time-aware attention can better summarize user preference than self-attention mechanism.} 
Our models with time-aware attention perform significantly better than the model with self-attention. For example, \textsc{EntityM-TimeA} achieves $1.9 \%$ higher Recall@1 than \textsc{EntityM-SelfA} with the same input (Context + TextEntity + DBPedia). This validates our intuition that the self-attention mechanism which lacks explicit guidance lead to sub-optimal recommendations and the recently appeared items are more important in reflecting user preference. It also shows the effectiveness of our designed time-aware attention. More analysis about it can be found in Section~\ref{sssec:para_effects1}.

$\bullet$~\textit{BART-based contextual-level representations are helpful.} We are the first to finetune a pre-trained BART model and utilize its representations for recommendation. As we can see in \Cref{tab:results_rec_redial}, the simplest \textsc{BART(ContextM)} model with only context as input achieves $35.0 \%$ Recall@50 while the \textsc{KBRD} model that incorporates DBPedia and text entities gets $33.6 \%$ Recall@50. We can also find that our models with time-aware attention show good improvements in all metrics after being enhanced with BART representations (i.e., with \textsc{ContextM}). Both indicate that contextual-level representations extracted by the finetuned BART models can reflect user preferences that can not be captured by entity-level representations, and so complement them well. \Cref{sfig:exp_bartrec_effect_recall} in Section \ref{sssec:exp:effective:contextual} shows more detailed analysis.

$\bullet$~\textit{Text entities are effective in capturing most relative items.} Our model \textsc{EntityM-TimeA} with additional text entities (i.e., TextEntity) as input can achieve better Recall@1 compared with the same model without text entities ($5.2 \%$ vs $4.6 \%$), while keeping similar Recall@10 and Recall@50. This means that text entities help re-rank the top predictions and find the most relative items for recommendation. 

\subsubsection{\textbf{Results on OpenDialKG}}\label{sec:exp_res:rec:opendial}
Apart from the ReDial that focuses on movie recommendation, we also examine our recommendation performance in a multi-domain dataset, OpenDialKG, to show the generalizability of our model. The results are displayed in \Cref{tab:results_rec_opendial}. Two observations can also be drawn as follows (We skip some of the same observations as Redial here).
\begin{table}[t]
\setlength{\tabcolsep}{0.5mm}
\caption{\label{tab:results_rec_opendial} \textbf{Recommendation} results (in \%) on OpenDialKG. 
}
\begin{center}
% \resizebox{\linewidth}{!}{
\begin{tabular}{l|ccccc}
\toprule[1.0pt]
\textbf{Models} & \textbf{Rec@1}  &\textbf{Rec@3} &\textbf{Rec@5} & \textbf{Rec@10} & \textbf{Rec@25} \\
\hline
\underline{Baselines} & & & & &  \\
seq2seq & 3.1 & 18.3 & 29.7 & 44.1 & 60.2 \\
Tri-LSTM & 3.2 & 14.2 & 22.6 & 36.3 & 56.2 \\
Ext-ED & 1.9 & 5.8 & 9.0 & 13.3 & 19.0 \\
DialKG Walker & 13.2 & 26.1 & 35.3 & 47.9 & 62.2 \\
\midrule[0.5pt]
\underline{Our Models} & & & &  &\\
\textsc{BART}(\textsc{ContextM}) & 5.8 & 19.7 & 31.0 & 45.5 & 57.8\\
\textsc{EntityM-SelfA} & 10.9 & 21.2 & 30.3 & 41.6 & 53.2 \\
\textsc{EntityM-TimeA} &16.0 &28.9 &34.3 &45.1 &57.9 \\
\textsc{EntityM-TimeA-ContextM} & \textbf{18.0} & \textbf{33.5} & \textbf{41.5} & \textbf{50.0} & \textbf{64.8} \\
\bottomrule[1.0pt]
\end{tabular}
% }
\end{center}
\end{table}
$\bullet$~\textit{The results on OpenDialKG are much higher than those on ReDial.}
As can be seen, our model achieves $18 \%$ Recall@1 score on OpenDialKG, which is much higher than $5.9 \%$ on ReDial. The reasons can be twofold. First, OpenDialKG is constructed based on the knowledge graph by some walking strategies~\cite{moon2019opendialkg}, which means that the item prediction would be easier when incorporating entity-level knowledge. Second, we can see that the conversations on OpenDialKG are much shorter than those on ReDial (see~\Cref{tab:statistic}). This indicates that it takes less turns leading to the final recommendation needed by the seekers and so the modeling of the appeared items on OpenDialKG may be simpler.

$\bullet$~\textit{Our model generalizes well to multiple domains.} As we can see from \Cref{tab:results_rec_opendial}, our model with time-aware attention and BART-enhanced representations (i.e., \textsc{EntityM-TimeA-ContextM}) achieves the best performance compared with all the other methods. We can observe similar trends among the different variants as those on ReDial, e.g., time-aware attention is better than self-attention and BART representations help improve all of the metrics. This validates that our method generalizes well to other domains.

\subsection{Generation Result Comparisons}
\label{sec:exp_res:gen}
In this subsection, we first discuss the automatic evaluation results in Section \ref{ssec:exp_res:gen:auto_eval} and then adopt a human evaluation to examine the generation results from a different perspective in Section \ref{ssec:exp_res:gen:human_eval}. 
\begin{table*}[t]
\caption{\label{tab:results_gen_redial} \textbf{Generation} results (in \%) on the ReDial dataset. ``BS'' refers to beam search, ``DBS'' refers to diverse beam search. 
% ``PT'' refers to pre-training.
% , and ``PPL'' refers to perplexity.
}
\begin{center}
% \resizebox{0.7\linewidth}{!}{
\begin{tabular}{l|ccc|cc|c}
\toprule[1.0pt]
\textbf{Models} & \textbf{Dist-2}  &\textbf{Dist-3} &\textbf{Dist-4} & \textbf{BLEU-2} & \textbf{BLEU-4} & \textbf{Perplexity$\downarrow$} \\
\midrule[0.5pt]
% \underline{Baselines} & & & & & &  \\
Transformer & 14.8 & 15.1 & 13.7 & - & - & - \\
ReDial & 22.5 & 23.6 & 22.8 & 17.8 & 7.4 & 61.7 \\
KBRD & 26.3 & 36.8 & 42.3 & 18.5 & 7.4 & 58.8 \\
KGSF & 28.9 & 43.4 & 51.9 & 16.4 & 7.4 & 131.1 \\
RevCore & 42.4 & 55.8 & 61.2 & - &- &- \\
\midrule[0.5pt]
Ours+BS &35.8 &49.9 &57.7 & \textbf{19.1} & \textbf{9.3} & 52.1 \\
$\;\;$ {\fontsize{8.5}{7.2}\selectfont - w/o BART Pre-train}&8.5 &10.9 &12.3 &18.6 &8.7 &\textbf{30.7} \\
Ours+DBS & \textbf{45.7} & \textbf{65.3} & \textbf{76.1} & \textbf{19.1} &8.9 & 54.8 \\
$\;\;$ {\fontsize{8.5}{7.2}\selectfont - w/o BART Pre-train}&13.9 &19.8 &23.8 &18.6 &8.2 &43.9 \\

\bottomrule[1.0pt]

\end{tabular}
% }
\end{center}
% \vskip -1em

% \vskip -1.5em
\end{table*}
\subsubsection{\textbf{Automatic Evaluation}}\label{ssec:exp_res:gen:auto_eval}
We show the automatic evaluation comparison results on ReDial in \Cref{tab:results_gen_redial}. To investigate the performance in different scenarios, we display the results of our models with conventional beam search (BS) and diverse beam search (DBS), respectively, together with their results without BART pre-training. Our model yields significantly better results on all of the evaluation metrics (including Dist-n, BLEU-n and perplexity) than the baselines, which indicates that our model tends to produce more diverse, coherent and fluent responses. In the following, we detail our observations:

$\bullet$~\textit{Our model is able to generate more diverse, coherent and fluent responses than the baselines.} As can be seen, our model with beam search achieves the best BLEU scores and perplexity, and our model with diverse beam search yields the highest Dist-n while maintaining comparable BLEU scores and perplexity.
Comparing our models with and without BART pre-training, the Dist-n metrics are affected severely, which validates the effectiveness of BART pre-training in enhancing the diversity of generated responses.

$\bullet$~\textit{It is challenging to balance between diversity and fluency.} 
The baselines tend to perform differently in terms of Dist-n and perplexity, e.g., KGSF achieves higher Dist-n than KBRD, but its perplexity is worse. 
We presume that higher diversity requires the models to extract more diverse patterns to express their content, but organizing them into a fluent response may be challenging. 
Another example is our model without BART pre-training achieves poor diversity (i.e., very low Dist-n scores), as it may overfit in the training corpus and tend to generate simple responses. But this also results in its lowest perplexity. Our model with diverse beam search achieves consistently better Dist-n scores, and maintains relatively lower perplexity, demonstrating the superiority of our model design.

$\bullet$~\textit{The Dist-n scores highly depend on the searching strategy.} Our model performs much better in terms of Dist-n when applying diverse beam search rather than conventional beam search. What is more, different configurations (e.g., length penalty in Section \ref{ssec:exp_res:ana:dist_n}) applied during response generation also affects the scores. This shows that it is not enough to evaluate the generation performance of these models only based on the metric, so we also include BLEU scores here and human evaluation (Section \ref{ssec:exp_res:gen:human_eval}) for better evaluation.

\subsubsection{\textbf{Human Evaluation}}\label{ssec:exp_res:gen:human_eval}
\begin{table}[t]
\caption{\label{tab:human_eval} \textbf{Human evaluation} of the \textbf{generation} results on the ReDial dataset. ``BS'' refers to beam search, ``DBS'' refers to diverse beam search. All the metrics are in the scale of [0, 2]. The overall Cohen's kappa coefficient is larger than 0.65.
}
\vskip -1em
\begin{center}
\resizebox{0.9\linewidth}{!}{\begin{tabular}{l|ccc}
\toprule[1.0pt]
\textbf{Models} & \textbf{Fluency}  & \textbf{Informativeness} & \textbf{Coherence} \\
\midrule[0.5pt]
HUMAN & \textbf{1.95} & \textbf{1.71}  & \textbf{1.71}  \\
\midrule[0.5pt]
ReDial & 1.92  &1.32 &1.23  \\
KBRD & \textbf{1.95} &1.39 &1.31 \\
KGSF & 1.91 & 1.02 & 0.95  \\
\midrule[0.5pt]
Ours+BS & \textbf{1.95} & 1.54 & 1.66 \\
Ours+DBS & 1.92 & 1.64 & 1.64 \\
\bottomrule[1.0pt]
\end{tabular}}

\end{center}
% \vskip -1em

% \vskip -1.0em
\end{table}
To conduct our human evaluation, we randomly sampled 100 context-response pairs from the test set and collected the corresponding generation results of our models as well as the baselines. We then employ two crowd-workers to score the results on the scale of [0, 1, 2], where higher scores indicate better quality. Following \cite{bao2020plato}, we evaluate the following three aspects of the results: 

\begin{itemize}[leftmargin=*,topsep=2pt,itemsep=2pt,parsep=0pt]
    \item \textbf{Fluency}: whether a response is in a proper English grammar and easy to understand.
    \item \textbf{Informativeness}: whether a response contains meaningful information. The ``safe responses'' are treated as uninformative as they may be repetitive and meaningless.
    \item \textbf{Coherence}: whether a response is coherent with its previous context, i.e., the discussed content should be consistent.
\end{itemize}
We display the human evaluation results in \Cref{tab:human_eval}. As can be seen, all the models generate responses with high fluency, but perform differently regarding informativeness and coherence. The baselines are more likely to produce safe responses (short and repetitive), resulting in lower informativeness. Also, their coherence scores are lower since they might produce some meaningless responses no matter what the previous contexts are.
Nevertheless, our model can generate more informative and coherent responses. 

\subsection{Effectiveness of the Proposed Mechanisms}\label{ssec:exp:effectiveness}
In this subsection, we first test the recommendation results with varying mentioned items in Section \ref{sssec:exp:effective:contextual} to show the effectiveness of the introduced contextual-level representation. Then we discuss the effects of BART pre-training for our model in Section \ref{sssec:exp:effective:pretrain}.

\subsubsection{\textbf{Effectiveness of Contextual-level Representation}}\label{sssec:exp:effective:contextual}
\begin{figure}[t]
\centering
\subfigure[Recall@50 (in \%) over Number of Mentioned Items]{\label{sfig:exp_bartrec_effect_recall}
\includegraphics[width=0.48\textwidth]{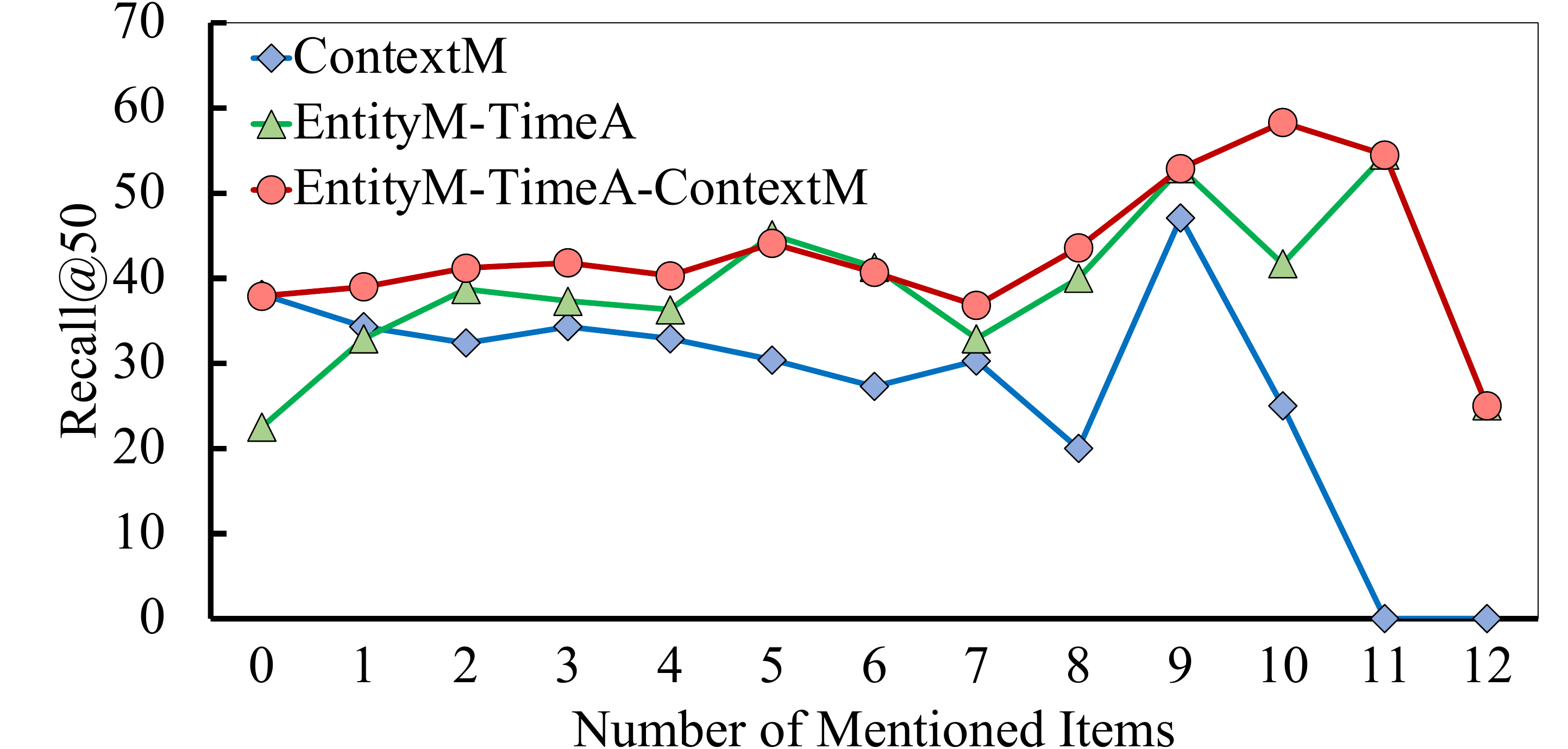}
}
% \vskip -0.5em
\subfigure[Number of cases over Number of Mentioned Items] {\label{sfig:exp_bartrec_effect_casenumber}
\includegraphics[width=0.48\textwidth]{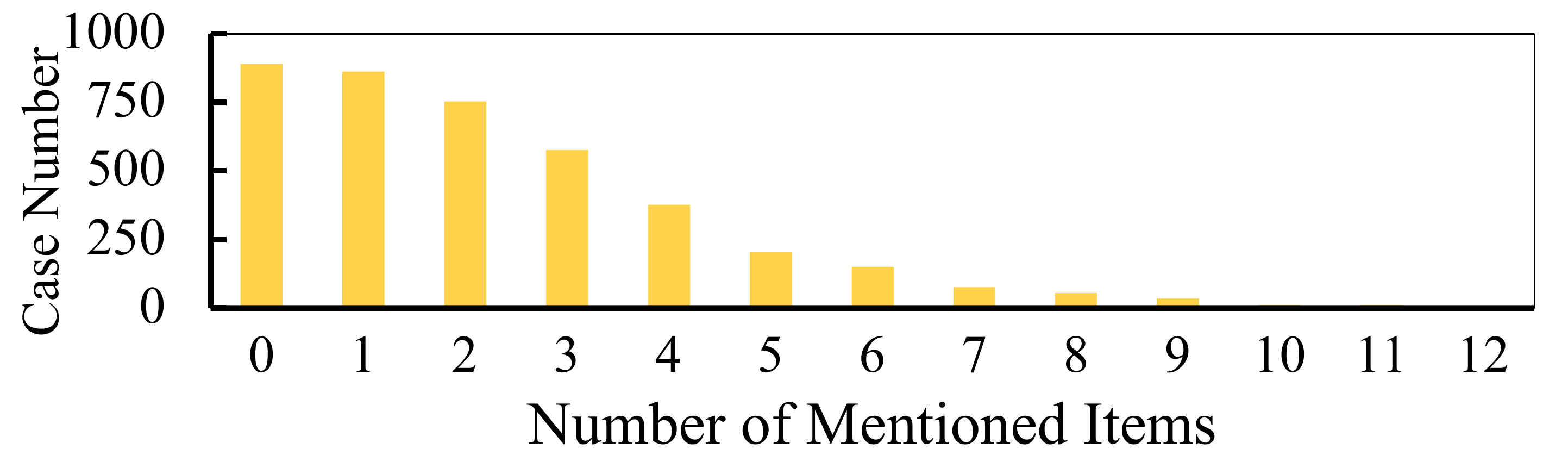}
}
\vskip -1em
\label{fig:exp_bartrec_effect}
\caption{Recall@50 of the models and number of cases over the number of mentioned items in context. 
}
% \vskip -1em
\end{figure}
\begin{table*}[t]\setlength{\tabcolsep}{2mm}
\caption{\label{tab:exp_rec_bart_pt} Recommendation results (in \%) of our models when using BART pre-training (PT) or not. The values in parentheses correspond to the percentage increase after adding PT.
}
\begin{center}
% \resizebox{\linewidth}{!}{
\begin{tabular}{l|l|c|lll}
\toprule[1.0pt]
\textbf{Models} & \textbf{Inputs}  & \textbf{Pre-training} & \textbf{R@1}  & \textbf{R@10} & \textbf{R@50} \\
% & & && &&
\midrule[0.5pt]
\multirow{2}{*}{\textsc{BART} (\textsc{ContextM})} & \multirow{2}{*}{Context} & \Checkmark  &3.0 (15.4\%) &16.4 (13.1\%)  &35.0 (10.4\%) \\ 
&&\XSolidBrush &2.6 &14.5  &31.7\\
\midrule[0.5pt]
\multirow{2}{*}{\textsc{EntityM-TimeA-ContextM}} & \multirow{2}{*}{Context+DBPedia} & \Checkmark  &5.7 (9.6\%)&22.6 (5.6\%)&40.4 (1.3\%) \\ 
&&\XSolidBrush &5.2&21.4&39.9\\
\midrule[0.5pt]
\multirow{2}{*}{\textsc{EntityM-TimeA-ContextM}} & \multirow{2}{*}{Context+TextEntity+DBPedia} & \Checkmark  &5.9 (7.3\%)  &24.0 (9.1\%) &41.3 (2.5\%) \\ 
&&\XSolidBrush &5.5  &22.0&40.3\\
\bottomrule[1.0pt]
\end{tabular}
% }
\end{center}
% \vskip -1em

% \vskip -1.5em
\end{table*}
To investigate how contextual-level representations influence recommendation, we show the Recall@50 scores with varying mentioned items numbers in context in \Cref{sfig:exp_bartrec_effect_recall} for our three model variants, \textsc{ContextM}, \textsc{EntityM-TimeA} and \textsc{EntityM-TimeA-ContextM}. The mentioned item number means how many items (e.g., movie, books) appear in the history utterances. Specifically, we divide the test samples into several groups based on the mentioned item number and calculate Recall@50 score for each group.

We can observe from \Cref{sfig:exp_bartrec_effect_recall} that the \textsc{ContextM} model performs better than \textsc{EntityM-TimeA} when the mentioned number is $0$ or $1$. 
Such a phenomenon is desired since when the item history information is rare or even missing, the entity-level representations are not sufficient to produce reliable recommendations while the contextual-level representations can capture useful information from the text. 
Then we find \textsc{EntityM-TimeA} performs consistently better than \textsc{ContextM} when the mentioned number increases. Because the increasing mentioned item number also means the context becomes longer, and the model might not be able to handle the long context, especially when the mentioned item number is larger than $10$. 
This means that contextual-level representations are useful in the \textbf{cold-start} scenario (not a rare situation as shown in \Cref{sfig:exp_bartrec_effect_casenumber}), which is a shortcoming for entity-level recommendation. Therefore, our full model (i.e., \textsc{EntityM-TimeA-ContextM}) combines the both representations and yields the best performance in all the situations.   

\subsubsection{\textbf{Effects of BART Pre-training}}\label{sssec:exp:effective:pretrain}
We elaborate the effects of BART pre-training (PT) here. For recommendation, we list the recall scores of our different model variants with and without BART PT in \Cref{tab:exp_rec_bart_pt}. As can be seen, all the models perform worse when removing pre-training, as pre-training on large-scale monolingual datasets helps the models learn better semantic features. When joint recommending with entity-level representations (i.e., \textsc{EntityM-TimeA-ContextM} model), the performance degradation becomes less, demonstrating the two kinds of information can complement each other. 

For generation, we have listed the ablation without pre-training in \Cref{tab:results_gen_redial}. 
As we can see, models without pre-training show poor performance on the Dist-n metrics as models tend to be overfitting and cannot generate diverse responses based on a small dataset. 

\subsection{Case Study}
\label{sec:exp_res:case}
\definecolor{mcolor1}{RGB}{41,18,171}
\begin{table}[t]
\caption{A conversation example on \textit{\textcolor{mcolor1}{movie item}} recommendation. The responses of \textbf{ReDial}, \textbf{KBRD} and \textbf{KGSF} are from the test results of the corresponding models. \textbf{Human} responses are the ground-truths from the corpus.}
% \vskip -1em
\centering
\begin{tabular}{rp{6.5cm}}
\toprule
% \multicolumn{2}{c}{{\bf Conversation Example}} \\
% \midrule
\textbf{Seeker}: & I have seen \textcolor{mcolor1}{The Conjuring (2013)} I very much enjoyed this movie. When I was younger I liked movies like \textcolor{mcolor1}{Friday the 13th (1980)}.  \\
\midrule
\textbf{Human}: & I liked that one too! \\
\textbf{ReDial}: & i have n't seen that one . i 've heard good things about it . \\
\textbf{KBRD}: & i have n't seen that one . i 've heard \textcolor{mcolor1}{Halloween (1978)} was good . \\
\textbf{KGSF}: & How about \textcolor{mcolor1}{The Descent (2005)} ? \\
\textbf{OUR}: & i haven't seen that one, but i've heard good things about it. \\
\midrule
\textbf{Seeker}: & I also liked to watch the \textcolor{mcolor1}{A Nightmare on Elm Street (1984)} Series. \\
\midrule
\textbf{Human}: & I loved that as well mainly because Johnny Depp is in it\\
\textbf{ReDial}: & i have n't seen that one yet . i 've seen \textcolor{mcolor1}{It  (2017)} \\
\textbf{KBRD}: & i have n't seen that one . i have heard \textcolor{mcolor1}{Halloween  (1978)} was good . \\
\textbf{KGSF}: & I also recommend \textcolor{mcolor1}{Insidious (2011)} \\
\textbf{OUR}: & I liked that one too! i also recommend \textcolor{mcolor1}{The Last Exorcism (2010)}.  \\
\midrule
\textbf{Seeker}: & I love Johnny Depp! So handsome, such a great actor. I love johnny Depp in \textcolor{mcolor1}{Alice in Wonderland (2010)}.  \\
\midrule
\textbf{Human}: &  Me too one of my favorites from him is \textcolor{mcolor1}{The Astronaut's Wife (1999)} \\
\textbf{ReDial}: & i have n't seen that one yet . i 've seen \textcolor{mcolor1}{It  (2017)} . \\
\textbf{KBRD}: & i have n't seen that one . i will check it out . \\
\textbf{KGSF}: & I hope you enjoy them !  \\
\textbf{OUR}: & Yes, I love Johnny Depp too ! Have you seen the new \textcolor{mcolor1}{It (2017)} movie? \\
\bottomrule
\end{tabular}
% \vskip -1em
\label{tab:case_study}
\end{table}
\begin{figure*}[th]
\centering
\subfigure[Effects of Time-Aware Parameter $\lambda$]{\label{sfig:three_paras:a}
\includegraphics[width=0.27\linewidth]{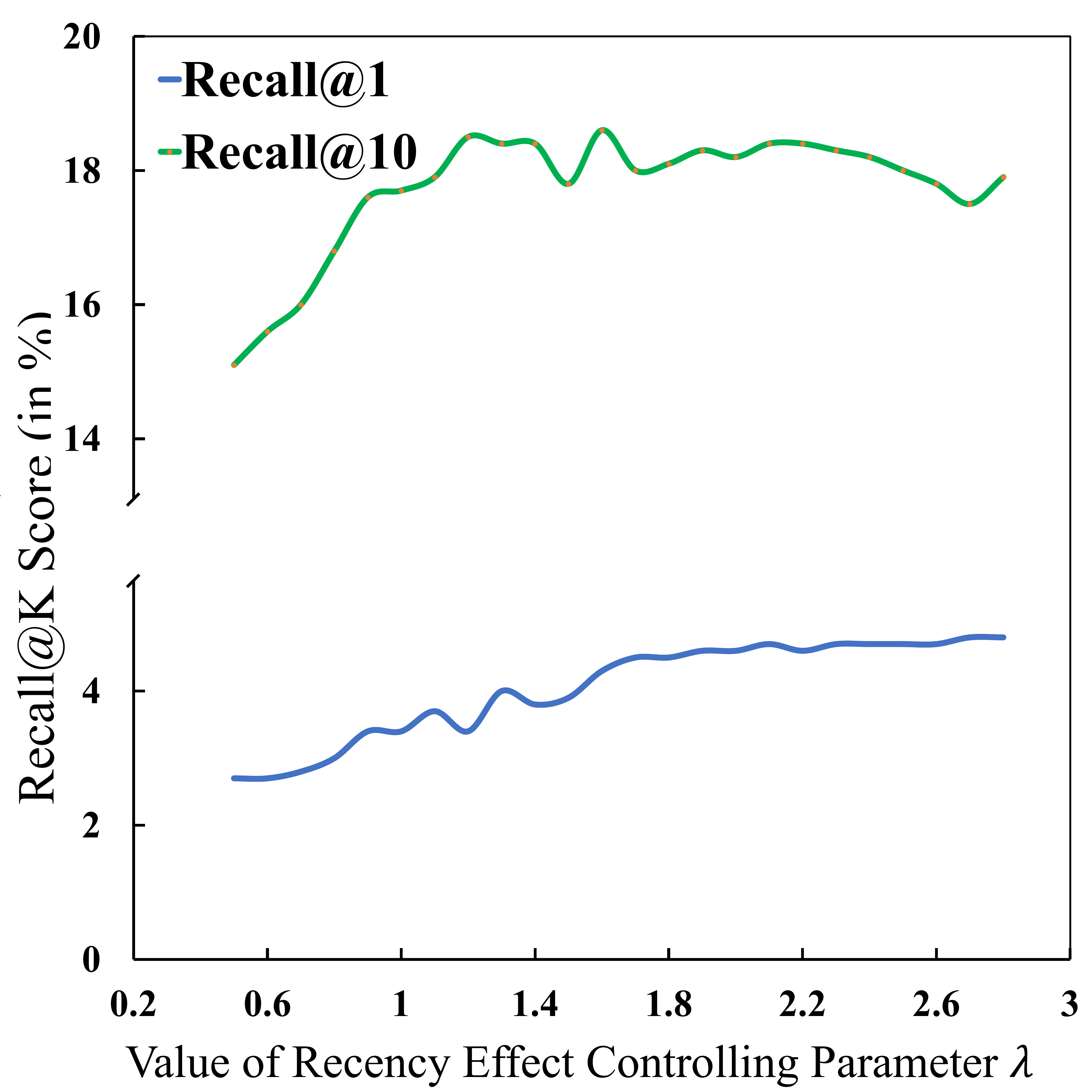}
}
\subfigure[Effects of Length Penalty ]{\label{sfig:three_paras:b}
\includegraphics[width=0.27\linewidth]{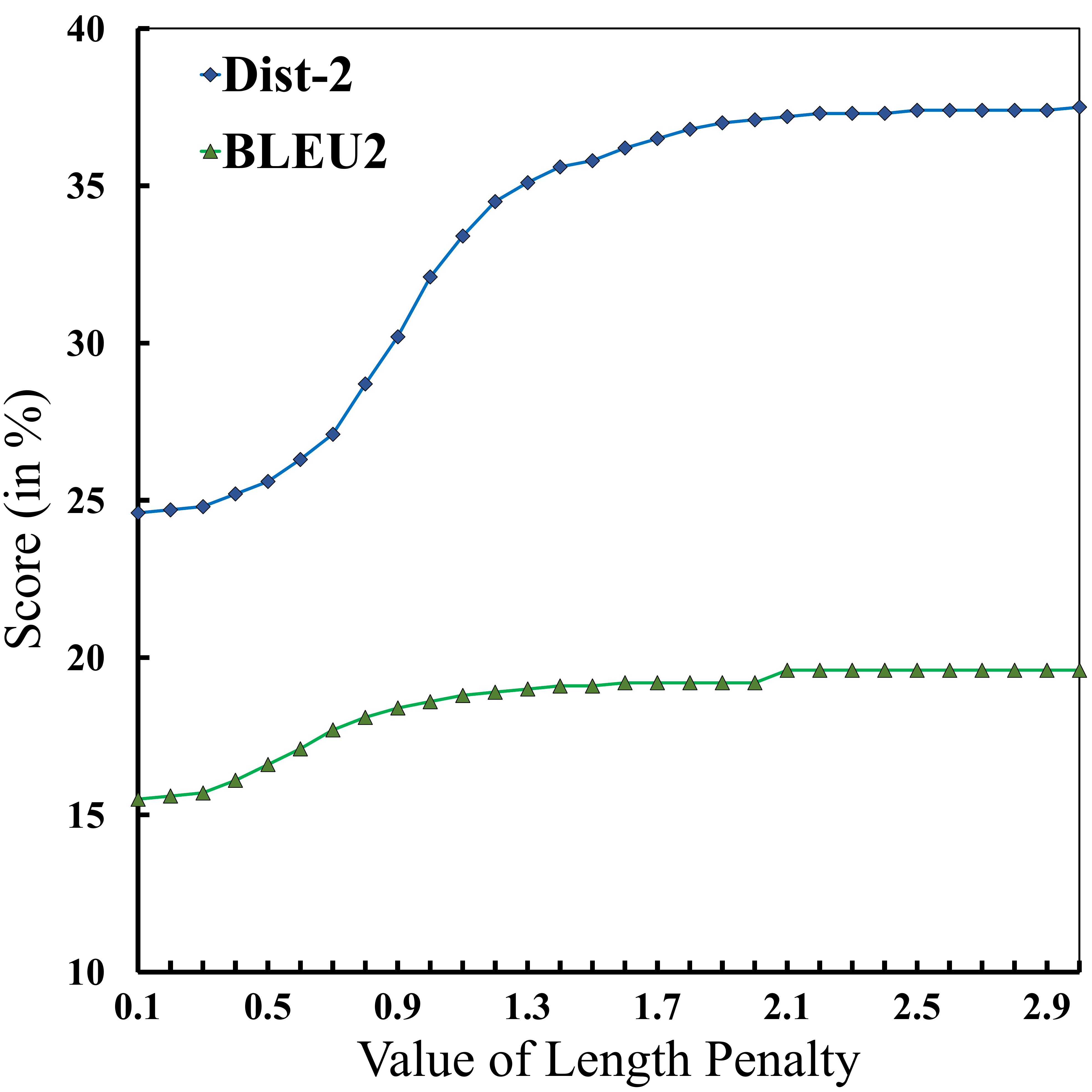}
}
\subfigure[Effects of the Trade-Off Parameter $\mu$ ]{\label{sfig:three_paras:c}
\includegraphics[width=0.27\linewidth]{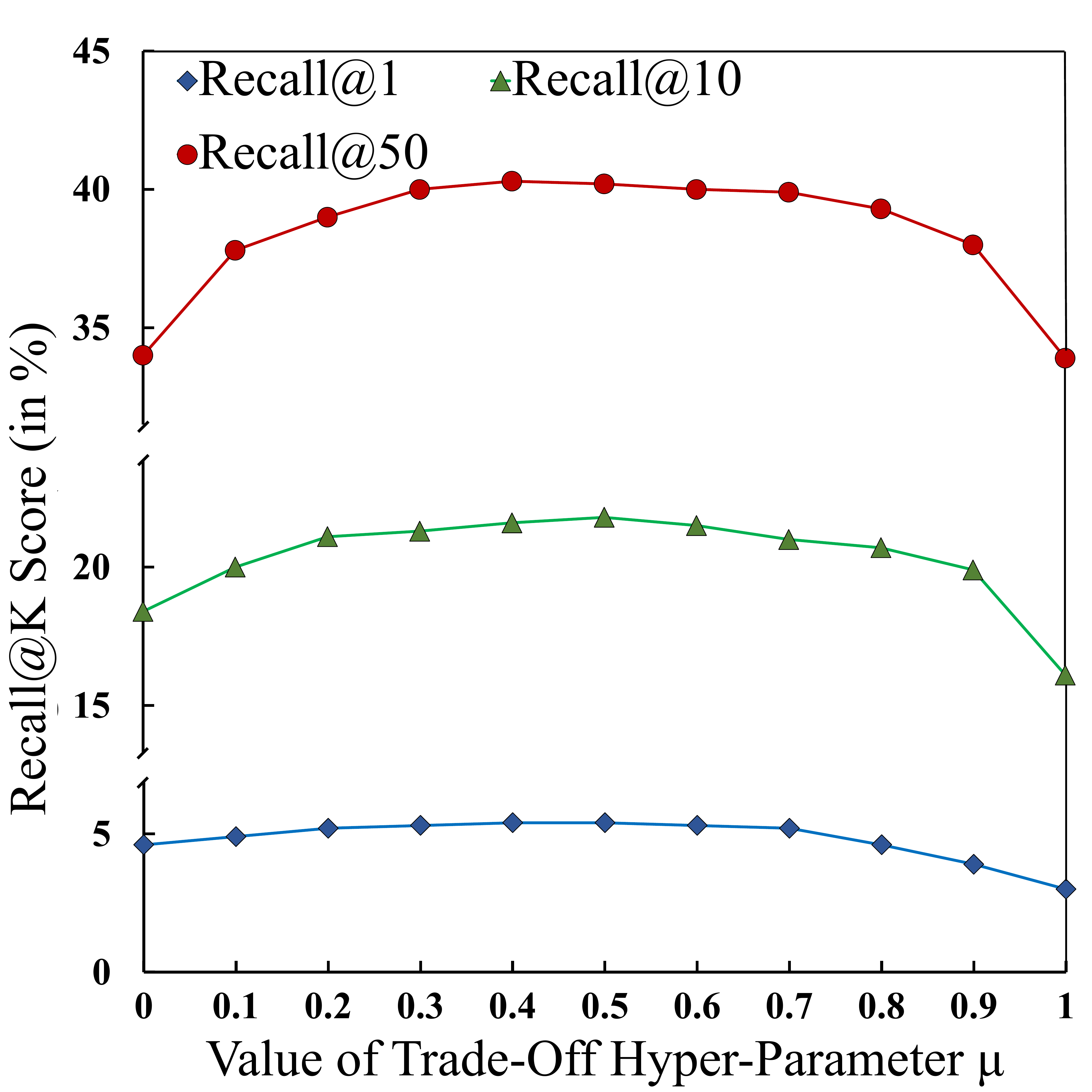}
}
\vskip -1.0em
\label{fig:three_paras}
\caption{Effects of three hyper-parameters: parameter $\lambda$ to control the time-aware user preference modeling, length penalty to influence the generation, and parameter $\mu$ to balance the effects of entity-level and contextual-level user representations.
}
% \vskip -1em
\end{figure*}

To give an intuitive assessment of the quality of the generated sentences and recommended items, we present a conversation example in \Cref{tab:case_study}. Apart from the results generated by our model and the human ground-truths from the corpus, we list the results of ReDial, KBRD and KGSF models. The reason why we do not list the results of RevCore and $C^2$-CRS is their released codes are not complete to re-implement the generation results. 

From \Cref{tab:case_study}, we can easily find that our model tends to generate responses with more diversity and coherence to the context, which may thank to the pre-trained BART model. For example, in the last turn, our model also talks about ``Johnny Depp'' which is mentioned in the turn of the seeker. This means that our model can capture the topic of discussion well while the baselines tend to repeat the boring responses with the same syntactic structure (e.g., ``I haven't seen that one....''). On the other hand, the baseline models like ReDial and KBRD tend to recommend the same items regardless of the history context, while our model can produce different recommendations based on different contexts thanks to our improved recommendation mechanism.

\subsection{Further Analysis}\label{ssec:further_analysis}
In this subsection, we extensively explore the effects of some hyper-parameters, together with in-depth discussions.

\subsubsection{\textbf{Influence of recency effect in Time-Aware Model}}\label{sssec:para_effects1}
We present the Recall@1 and Recall@10 (Recall@50 shows a similar trend with Recall@10 so we omit here) scores with varying values of $\lambda$ in \Cref{sfig:three_paras:a}, which controls the effect of recency introduced in our proposed Time-Aware Attention (see \Cref{eq:time}). $\lambda < 1$ indicates the earlier appeared items are more important, which results in quick performance drops. 
And the performance increases when $\lambda > 1$ compared to $\lambda = 1$. This validates the intuition that more recently appeared items contribute more to the next item recommendation. On the other hand, Recall@1 and Recall@10 present different trends when $\lambda$ increases -- Recall@1 consistently increases while Recall@10 begins to drop when $\lambda > 2$. This is because when $\lambda$ is too large, time-aware attention tends to concern with the most recently appeared item. This is helpful in finding the most relative items but hurts the overall recommendation.

\subsubsection{\textbf{Limitation of Dist-n Metrics.}}\label{sssec:para_effects2}
\label{ssec:exp_res:ana:dist_n}
As we find that search strategies seriously affect the Dist-n metrics (see Section \ref{ssec:exp_res:gen:auto_eval}), we present more analysis on them by exploring the length penalty (a hyper-parameter that can control the lengths of final generated results) when generating the responses. We display the results of Dist-2 and BLEU-2 for our \textsc{EntityM-TimeA-ContextM} model with different length penalty in \Cref{sfig:three_paras:b} (other metrics are in similar trends). We can find that the generated lengths also affect a lot the Dist-n scores since longer responses allow more different tokens to be generated. However, this is not expected as the longer is not necessarily the better. Therefore, other metrics, including human evaluation, are desired to explicitly evaluate generation performance.

\subsubsection{\textbf{Trade-off between Entity-level and Contextual-level Representations.}}\label{sssec:para_effects3}
We examine the effects of the trade-off hyper-parameter $\mu$ in \Cref{eq:rec-joint} by setting its value from 0 (only entity-level representations) to 1 (only contextual-level representations) and display the results of our \textsc{EntityM-TimeA-ContextM} model in \Cref{sfig:three_paras:c}. As can be seen, Recall@50 score is significantly improved when the value of $\mu$ changes from 0 to 0.1 (or 1 to 0.9). This validates that the two representations can capture user preferences from different perspectives and complement each other for better recommendation performance. Besides, the trends of Recall@1 and Recall@10 are similar to that of Recall@50. The best results are achieved around $\mu=0.5$, showing that both levels of representation are important.

\section{Conclusion and Future Work}\label{sec:conlusion}
In this work, we focus on conversational recommender systems and propose to capture both entity-level and contextual-level representations to improve the recommendation performance, where a time-aware user preference modeling is designed to better capture the user's interests and a pre-trained BART model is used to enhance contextual-level user preference modeling and improve the diversity of the generated responses. Experiments on two publicly available datasets show that the proposed model can achieve better performance with less external domain-specific knowledge and generalizes well to other domains. Further analyses also examine the effectiveness of our model in different scenarios.

In future work, we would like to explore more accurate user preference modeling mechanisms. Though our current time-aware modeling method shows promising improvement on both datasets, it is based on a simple assumption and cannot deal with more complicated real-world scenarios. A combined mechanism that concerns time effects and other features like sentiment is desired. However, it is hard to train a powerful user preference learning framework with the currently available datasets, which are human-created and relatively small. A large-scale, diverse, and multi-domain dataset might be needed in the future.

\ifCLASSOPTIONcaptionsoff
  \newpage
\fi



\bibliographystyle{IEEEtran}
\bibliography{anthology,custom}

\begin{thebibliography}{10}
\providecommand{\url}[1]{#1}
\csname url@samestyle\endcsname
\providecommand{\newblock}{\relax}
\providecommand{\bibinfo}[2]{#2}
\providecommand{\BIBentrySTDinterwordspacing}{\spaceskip=0pt\relax}
\providecommand{\BIBentryALTinterwordstretchfactor}{4}
\providecommand{\BIBentryALTinterwordspacing}{\spaceskip=\fontdimen2\font plus
\BIBentryALTinterwordstretchfactor\fontdimen3\font minus
  \fontdimen4\font\relax}
\providecommand{\BIBforeignlanguage}[2]{{%
\expandafter\ifx\csname l@#1\endcsname\relax
\typeout{** WARNING: IEEEtran.bst: No hyphenation pattern has been}%
\typeout{** loaded for the language `#1'. Using the pattern for}%
\typeout{** the default language instead.}%
\else
\language=\csname l@#1\endcsname
\fi
#2}}
\providecommand{\BIBdecl}{\relax}
\BIBdecl

\bibitem{li2018towards}
R.~Li, S.~Kahou, H.~Schulz, V.~Michalski, L.~Charlin, and C.~Pal, ``Towards
  deep conversational recommendations,'' \emph{arXiv preprint
  arXiv:1812.07617}, 2018.

\bibitem{chen-etal-2019-towards}
\BIBentryALTinterwordspacing
Q.~Chen, J.~Lin, Y.~Zhang, M.~Ding, Y.~Cen, H.~Yang, and J.~Tang, ``Towards
  knowledge-based recommender dialog system,'' in \emph{Proceedings of the 2019
  Conference on Empirical Methods in Natural Language Processing and the 9th
  International Joint Conference on Natural Language Processing
  (EMNLP-IJCNLP)}.\hskip 1em plus 0.5em minus 0.4em\relax Hong Kong, China:
  Association for Computational Linguistics, Nov. 2019, pp. 1803--1813.
  [Online]. Available: \url{https://aclanthology.org/D19-1189}
\BIBentrySTDinterwordspacing

\bibitem{zhou2020improving}
K.~Zhou, W.~X. Zhao, S.~Bian, Y.~Zhou, J.-R. Wen, and J.~Yu, ``Improving
  conversational recommender systems via knowledge graph based semantic
  fusion,'' in \emph{Proceedings of the 26th ACM SIGKDD}, 2020, pp. 1006--1014.

\bibitem{lu2021revcore}
Y.~Lu, J.~Bao, Y.~Song, Z.~Ma, S.~Cui, Y.~Wu, and X.~He, ``Revcore:
  Review-augmented conversational recommendation,'' \emph{arXiv preprint
  arXiv:2106.00957}, 2021.

\bibitem{moon2019opendialkg}
S.~Moon, P.~Shah, A.~Kumar, and R.~Subba, ``Opendialkg: Explainable
  conversational reasoning with attention-based walks over knowledge graphs,''
  in \emph{Proceedings of the 57th ACL}, 2019, pp. 845--854.

\bibitem{lewis-etal-2020-bart}
\BIBentryALTinterwordspacing
M.~Lewis, Y.~Liu, N.~Goyal, M.~Ghazvininejad, A.~Mohamed, O.~Levy, V.~Stoyanov,
  and L.~Zettlemoyer, ``{BART}: Denoising sequence-to-sequence pre-training for
  natural language generation, translation, and comprehension,'' in
  \emph{Proceedings of the 58th Annual Meeting of the Association for
  Computational Linguistics}.\hskip 1em plus 0.5em minus 0.4em\relax Online:
  Association for Computational Linguistics, Jul. 2020, pp. 7871--7880.
  [Online]. Available: \url{https://aclanthology.org/2020.acl-main.703}
\BIBentrySTDinterwordspacing

\bibitem{wang2021finetuning}
L.~Wang, H.~Hu, L.~Sha, C.~Xu, K.-F. Wong, and D.~Jiang, ``Finetuning
  large-scale pre-trained language models for conversational recommendation
  with knowledge graph,'' \emph{arXiv preprint arXiv:2110.07477}, 2021.

\bibitem{christakopoulou2018q}
K.~Christakopoulou, A.~Beutel, R.~Li, S.~Jain, and E.~H. Chi, ``Q\&r: A
  two-stage approach toward interactive recommendation,'' in \emph{Proceedings
  of the 24th ACM SIGKDD}, 2018, pp. 139--148.

\bibitem{zhang2018towards}
Y.~Zhang, X.~Chen, Q.~Ai, L.~Yang, and W.~B. Croft, ``Towards conversational
  search and recommendation: System ask, user respond,'' in \emph{Proceedings
  of the 27th acm international conference on information and knowledge
  management}, 2018, pp. 177--186.

\bibitem{aliannejadi2019asking}
M.~Aliannejadi, H.~Zamani, F.~Crestani, and W.~B. Croft, ``Asking clarifying
  questions in open-domain information-seeking conversations,'' in
  \emph{Proceedings of the 42nd international acm sigir conference}, 2019, pp.
  475--484.

\bibitem{lei2020estimation}
W.~Lei, X.~He, Y.~Miao, Q.~Wu, R.~Hong, M.-Y. Kan, and T.-S. Chua,
  ``Estimation-action-reflection: Towards deep interaction between
  conversational and recommender systems,'' in \emph{Proceedings of the 13th
  International Conference on Web Search and Data Mining}, 2020, pp. 304--312.

\bibitem{ren2020crsal}
X.~Ren, H.~Yin, T.~Chen, H.~Wang, N.~Q.~V. Hung, Z.~Huang, and X.~Zhang,
  ``Crsal: Conversational recommender systems with adversarial learning,''
  \emph{ACM Transactions on Information Systems (TOIS)}, vol.~38, no.~4, pp.
  1--40, 2020.

\bibitem{deng2021unified}
Y.~Deng, Y.~Li, F.~Sun, B.~Ding, and W.~Lam, ``Unified conversational
  recommendation policy learning via graph-based reinforcement learning,''
  \emph{arXiv preprint arXiv:2105.09710}, 2021.

\bibitem{xu-etal-2020-user}
\BIBentryALTinterwordspacing
H.~Xu, S.~Moon, H.~Liu, B.~Liu, P.~Shah, B.~Liu, and P.~Yu, ``User memory
  reasoning for conversational recommendation,'' in \emph{Proceedings of the
  28th International Conference on Computational Linguistics}.\hskip 1em plus
  0.5em minus 0.4em\relax Barcelona, Spain (Online): International Committee on
  Computational Linguistics, Dec. 2020, pp. 5288--5308. [Online]. Available:
  \url{https://aclanthology.org/2020.coling-main.463}
\BIBentrySTDinterwordspacing

\bibitem{lei2020interactive}
W.~Lei, G.~Zhang, X.~He, Y.~Miao, X.~Wang, L.~Chen, and T.~Chua, ``Interactive
  path reasoning on graph for conversational recommendation,'' in \emph{{KDD}
  '20: The 26th {ACM} {SIGKDD} Conference on Knowledge Discovery and Data
  Mining, Virtual Event, CA, USA, August 23-27, 2020}, R.~Gupta, Y.~Liu,
  J.~Tang, and B.~A. Prakash, Eds.\hskip 1em plus 0.5em minus 0.4em\relax
  {ACM}, 2020, pp. 2073--2083.

\bibitem{ren2021learning}
X.~Ren, H.~Yin, T.~Chen, H.~Wang, Z.~Huang, and K.~Zheng, ``Learning to ask
  appropriate questions in conversational recommendation,'' \emph{arXiv
  preprint arXiv:2105.04774}, 2021.

\bibitem{xu2021adapting}
K.~Xu, J.~Yang, J.~Xu, S.~Gao, J.~Guo, and J.-R. Wen, ``Adapting user
  preference to online feedback in multi-round conversational recommendation,''
  in \emph{Proceedings of the 14th ACM International Conference on Web Search
  and Data Mining}, 2021, pp. 364--372.

\bibitem{li2010contextual}
L.~Li, W.~Chu, J.~Langford, and R.~E. Schapire, ``A contextual-bandit approach
  to personalized news article recommendation,'' in \emph{Proceedings of the
  19th WWW}, 2010, pp. 661--670.

\bibitem{li2016collaborative}
S.~Li, A.~Karatzoglou, and C.~Gentile, ``Collaborative filtering bandits,'' in
  \emph{Proceedings of the 39th International ACM SIGIR}, 2016, pp. 539--548.

\bibitem{christakopoulou2016towards}
K.~Christakopoulou, F.~Radlinski, and K.~Hofmann, ``Towards conversational
  recommender systems,'' in \emph{Proceedings of the 22nd ACM SIGKDD}, 2016,
  pp. 815--824.

\bibitem{li2020seamlessly}
S.~Li, W.~Lei, Q.~Wu, X.~He, P.~Jiang, and T.-S. Chua, ``Seamlessly unifying
  attributes and items: Conversational recommendation for cold-start users,''
  \emph{arXiv preprint arXiv:2005.12979}, 2020.

\bibitem{sun2018conversational}
Y.~Sun and Y.~Zhang, ``Conversational recommender system,'' in \emph{The 41st
  international acm sigir conference on research \& development in information
  retrieval}, 2018, pp. 235--244.

\bibitem{zhou2020towards}
K.~Zhou, Y.~Zhou, W.~X. Zhao, X.~Wang, and J.-R. Wen, ``Towards topic-guided
  conversational recommender system,'' \emph{arXiv preprint arXiv:2010.04125},
  2020.

\bibitem{kang2019recommendation}
D.~Kang, A.~Balakrishnan, P.~Shah, P.~Crook, Y.-L. Boureau, and J.~Weston,
  ``Recommendation as a communication game: Self-supervised bot-play for
  goal-oriented dialogue,'' \emph{arXiv preprint arXiv:1909.03922}, 2019.

\bibitem{liu2020towards}
Z.~Liu, H.~Wang, Z.-Y. Niu, H.~Wu, W.~Che, and T.~Liu, ``Towards conversational
  recommendation over multi-type dialogs,'' \emph{arXiv preprint
  arXiv:2005.03954}, 2020.

\bibitem{hayati2020inspired}
S.~A. Hayati, D.~Kang, Q.~Zhu, W.~Shi, and Z.~Yu, ``Inspired: Toward sociable
  recommendation dialog systems,'' \emph{arXiv preprint arXiv:2009.14306},
  2020.

\bibitem{batmaz2019review}
Z.~Batmaz, A.~Yurekli, A.~Bilge, and C.~Kaleli, ``A review on deep learning for
  recommender systems: challenges and remedies,'' \emph{Artificial Intelligence
  Review}, vol.~52, no.~1, pp. 1--37, 2019.

\bibitem{hu2008collaborative}
Y.~Hu, Y.~Koren, and C.~Volinsky, ``Collaborative filtering for implicit
  feedback datasets,'' in \emph{2008 Eighth IEEE international conference on
  data mining}.\hskip 1em plus 0.5em minus 0.4em\relax Ieee, 2008, pp.
  263--272.

\bibitem{schafer2007collaborative}
J.~B. Schafer, D.~Frankowski, J.~Herlocker, and S.~Sen, ``Collaborative
  filtering recommender systems,'' in \emph{The adaptive web}.\hskip 1em plus
  0.5em minus 0.4em\relax Springer, 2007, pp. 291--324.

\bibitem{van2000using}
R.~Van~Meteren and M.~Van~Someren, ``Using content-based filtering for
  recommendation,'' in \emph{Proceedings of the machine learning in the new
  information age: MLnet/ECML2000 workshop}, vol.~30, 2000, pp. 47--56.

\bibitem{tran2000hybrid}
T.~Tran and R.~Cohen, ``Hybrid recommender systems for electronic commerce,''
  in \emph{Proc. Knowledge-Based Electronic Markets, Papers from the AAAI
  Workshop, Technical Report WS-00-04, AAAI Press}, vol.~40, 2000.

\bibitem{sarwar2001item}
B.~Sarwar, G.~Karypis, J.~Konstan, and J.~Riedl, ``Item-based collaborative
  filtering recommendation algorithms,'' in \emph{Proceedings of the 10th
  international conference on World Wide Web}, 2001, pp. 285--295.

\bibitem{breese2013empirical}
J.~S. Breese, D.~Heckerman, and C.~Kadie, ``Empirical analysis of predictive
  algorithms for collaborative filtering,'' \emph{arXiv preprint
  arXiv:1301.7363}, 2013.

\bibitem{ungar1998clustering}
L.~H. Ungar and D.~P. Foster, ``Clustering methods for collaborative
  filtering,'' in \emph{AAAI workshop on recommendation systems}, vol.~1.\hskip
  1em plus 0.5em minus 0.4em\relax Menlo Park, CA, 1998, pp. 114--129.

\bibitem{su2006collaborative}
X.~Su and T.~M. Khoshgoftaar, ``Collaborative filtering for multi-class data
  using belief nets algorithms,'' in \emph{2006 18th IEEE international
  conference on Tools with Artificial Intelligence (ICTAI'06)}.\hskip 1em plus
  0.5em minus 0.4em\relax IEEE, 2006, pp. 497--504.

\bibitem{koren2009matrix}
Y.~Koren, R.~Bell, and C.~Volinsky, ``Matrix factorization techniques for
  recommender systems,'' \emph{Computer}, vol.~42, no.~8, pp. 30--37, 2009.

\bibitem{lops2011content}
P.~Lops, M.~d. Gemmis, and G.~Semeraro, ``Content-based recommender systems:
  State of the art and trends,'' \emph{Recommender systems handbook}, pp.
  73--105, 2011.

\bibitem{devlin-etal-2019-bert}
\BIBentryALTinterwordspacing
J.~Devlin, M.-W. Chang, K.~Lee, and K.~Toutanova, ``{BERT}: Pre-training of
  deep bidirectional transformers for language understanding,'' in
  \emph{Proceedings of the 2019 Conference of the North {A}merican Chapter of
  the Association for Computational Linguistics: Human Language Technologies,
  Volume 1 (Long and Short Papers)}.\hskip 1em plus 0.5em minus 0.4em\relax
  Minneapolis, Minnesota: Association for Computational Linguistics, Jun. 2019,
  pp. 4171--4186. [Online]. Available: \url{https://aclanthology.org/N19-1423}
\BIBentrySTDinterwordspacing

\bibitem{liu2019roberta}
Y.~Liu, M.~Ott, N.~Goyal, J.~Du, M.~Joshi, D.~Chen, O.~Levy, M.~Lewis,
  L.~Zettlemoyer, and V.~Stoyanov, ``Roberta: A robustly optimized bert
  pretraining approach,'' \emph{arXiv preprint arXiv:1907.11692}, 2019.

\bibitem{brown2020language}
T.~Brown, B.~Mann, N.~Ryder, M.~Subbiah, J.~D. Kaplan, P.~Dhariwal,
  A.~Neelakantan, P.~Shyam, G.~Sastry, A.~Askell \emph{et~al.}, ``Language
  models are few-shot learners,'' \emph{Advances in neural information
  processing systems}, vol.~33, pp. 1877--1901, 2020.

\bibitem{fedus2021switch}
W.~Fedus, B.~Zoph, and N.~Shazeer, ``Switch transformers: Scaling to trillion
  parameter models with simple and efficient sparsity,'' \emph{arXiv preprint
  arXiv:2101.03961}, 2021.

\bibitem{mikolov2013distributed}
T.~Mikolov, I.~Sutskever, K.~Chen, G.~S. Corrado, and J.~Dean, ``Distributed
  representations of words and phrases and their compositionality,'' in
  \emph{Advances in neural information processing systems}, 2013, pp.
  3111--3119.

\bibitem{bengio2003neural}
Y.~Bengio, R.~Ducharme, P.~Vincent, and C.~Jauvin, ``A neural probabilistic
  language model,'' \emph{Journal of machine learning research}, vol.~3, no.
  Feb, pp. 1137--1155, 2003.

\bibitem{wang2015unsupervised}
X.~Wang and A.~Gupta, ``Unsupervised learning of visual representations using
  videos,'' in \emph{Proceedings of the IEEE international conference on
  computer vision}, 2015, pp. 2794--2802.

\bibitem{petroni-etal-2019-language}
\BIBentryALTinterwordspacing
F.~Petroni, T.~Rockt{\"a}schel, S.~Riedel, P.~Lewis, A.~Bakhtin, Y.~Wu, and
  A.~Miller, ``Language models as knowledge bases?'' in \emph{Proceedings of
  the 2019 Conference on Empirical Methods in Natural Language Processing and
  the 9th International Joint Conference on Natural Language Processing
  (EMNLP-IJCNLP)}.\hskip 1em plus 0.5em minus 0.4em\relax Hong Kong, China:
  Association for Computational Linguistics, Nov. 2019, pp. 2463--2473.
  [Online]. Available: \url{https://aclanthology.org/D19-1250}
\BIBentrySTDinterwordspacing

\bibitem{gao2020making}
T.~Gao, A.~Fisch, and D.~Chen, ``Making pre-trained language models better
  few-shot learners,'' \emph{arXiv preprint arXiv:2012.15723}, 2020.

\bibitem{liu2021gpt}
X.~Liu, Y.~Zheng, Z.~Du, M.~Ding, Y.~Qian, Z.~Yang, and J.~Tang, ``Gpt
  understands, too,'' \emph{arXiv preprint arXiv:2103.10385}, 2021.

\bibitem{zhou2022c2}
Y.~Zhou, K.~Zhou, W.~X. Zhao, C.~Wang, P.~Jiang, and H.~Hu, ``C2-crs:
  Coarse-to-fine contrastive learning for conversational recommender system,''
  \emph{arXiv preprint arXiv:2201.02732}, 2022.

\bibitem{DBLP:conf/nips/VaswaniSPUJGKP17}
A.~Vaswani, N.~Shazeer, N.~Parmar, J.~Uszkoreit, L.~Jones, A.~N. Gomez,
  L.~Kaiser, and I.~Polosukhin, ``Attention is all you need,'' in
  \emph{Advances in Neural Information Processing Systems 30: Annual Conference
  on Neural Information Processing Systems 2017, December 4-9, 2017, Long
  Beach, CA, {USA}}, I.~Guyon, U.~von Luxburg, S.~Bengio, H.~M. Wallach,
  R.~Fergus, S.~V.~N. Vishwanathan, and R.~Garnett, Eds., 2017, pp. 5998--6008.

\bibitem{daiber2013improving}
J.~Daiber, M.~Jakob, C.~Hokamp, and P.~N. Mendes, ``Improving efficiency and
  accuracy in multilingual entity extraction,'' in \emph{Proceedings of the 9th
  international conference on semantic systems}, 2013, pp. 121--124.

\bibitem{schlichtkrull2018modeling}
M.~Schlichtkrull, T.~N. Kipf, P.~Bloem, R.~Van Den~Berg, I.~Titov, and
  M.~Welling, ``Modeling relational data with graph convolutional networks,''
  in \emph{European semantic web conference}.\hskip 1em plus 0.5em minus
  0.4em\relax Springer, 2018, pp. 593--607.

\bibitem{gu-etal-2016-incorporating}
\BIBentryALTinterwordspacing
J.~Gu, Z.~Lu, H.~Li, and V.~O. Li, ``Incorporating copying mechanism in
  sequence-to-sequence learning,'' in \emph{Proceedings of the 54th Annual
  Meeting of the Association for Computational Linguistics (Volume 1: Long
  Papers)}.\hskip 1em plus 0.5em minus 0.4em\relax Berlin, Germany: Association
  for Computational Linguistics, Aug. 2016, pp. 1631--1640. [Online].
  Available: \url{https://aclanthology.org/P16-1154}
\BIBentrySTDinterwordspacing

\bibitem{gulcehre-etal-2016-pointing}
\BIBentryALTinterwordspacing
C.~Gulcehre, S.~Ahn, R.~Nallapati, B.~Zhou, and Y.~Bengio, ``Pointing the
  unknown words,'' in \emph{Proceedings of the 54th Annual Meeting of the
  Association for Computational Linguistics (Volume 1: Long Papers)}.\hskip 1em
  plus 0.5em minus 0.4em\relax Berlin, Germany: Association for Computational
  Linguistics, Aug. 2016, pp. 140--149. [Online]. Available:
  \url{https://aclanthology.org/P16-1014}
\BIBentrySTDinterwordspacing

\bibitem{sordoni2015hierarchical}
A.~Sordoni, Y.~Bengio, H.~Vahabi, C.~Lioma, J.~Grue~Simonsen, and J.-Y. Nie,
  ``A hierarchical recurrent encoder-decoder for generative context-aware query
  suggestion,'' in \emph{proceedings of the 24th ACM international on
  conference on information and knowledge management}, 2015, pp. 553--562.

\bibitem{he2017distributed}
J.~He, H.~H. Zhuo, and J.~Law, ``Distributed-representation based hybrid
  recommender system with short item descriptions,'' \emph{arXiv preprint
  arXiv:1703.04854}, 2017.

\bibitem{ma2020bridging}
W.~Ma, R.~Takanobu, M.~Tu, and M.~Huang, ``Bridging the gap between
  conversational reasoning and interactive recommendation,'' \emph{arXiv
  preprint arXiv:2010.10333}, 2020.

\bibitem{viola1997alignment}
P.~Viola and W.~M. Wells~III, ``Alignment by maximization of mutual
  information,'' \emph{International journal of computer vision}, vol.~24,
  no.~2, pp. 137--154, 1997.

\bibitem{sutskever2014sequence}
I.~Sutskever, O.~Vinyals, and Q.~V. Le, ``Sequence to sequence learning with
  neural networks,'' \emph{Advances in neural information processing systems},
  vol.~27, 2014.

\bibitem{young1709augmenting}
T.~Young, E.~Cambria, I.~Chaturvedi, M.~Huang, H.~Zhou, and S.~Biswas,
  ``Augmenting end-to-end dialog systems with commonsense knowledge (2017),''
  \emph{arXiv preprint arXiv:1709.05453}.

\bibitem{parthasarathi2018extending}
P.~Parthasarathi and J.~Pineau, ``Extending neural generative conversational
  model using external knowledge sources,'' \emph{arXiv preprint
  arXiv:1809.05524}, 2018.

\bibitem{vijayakumar2016diverse}
A.~K. Vijayakumar, M.~Cogswell, R.~R. Selvaraju, Q.~Sun, S.~Lee, D.~Crandall,
  and D.~Batra, ``Diverse beam search: Decoding diverse solutions from neural
  sequence models,'' \emph{arXiv preprint arXiv:1610.02424}, 2016.

\bibitem{ott2019fairseq}
M.~Ott, S.~Edunov, A.~Baevski, A.~Fan, S.~Gross, N.~Ng, D.~Grangier, and
  M.~Auli, ``fairseq: A fast, extensible toolkit for sequence modeling,''
  \emph{arXiv preprint arXiv:1904.01038}, 2019.

\bibitem{bao2020plato}
S.~Bao, H.~He, F.~Wang, H.~Wu, H.~Wang, W.~Wu, Z.~Guo, Z.~Liu, and X.~Xu,
  ``Plato-2: Towards building an open-domain chatbot via curriculum learning,''
  \emph{arXiv preprint arXiv:2006.16779}, 2020.

\end{thebibliography}


% % \bibliography{anthology,custom}
% \end{thebibliography}
% \begin{IEEEbiography}[{\includegraphics[width=1in,height=1.25in,clip,keepaspectratio]{mshell}}]{Michael Shell}
% \newpage
\begin{IEEEbiography}[{\includegraphics[width=1in,height=1.25in,clip,keepaspectratio]{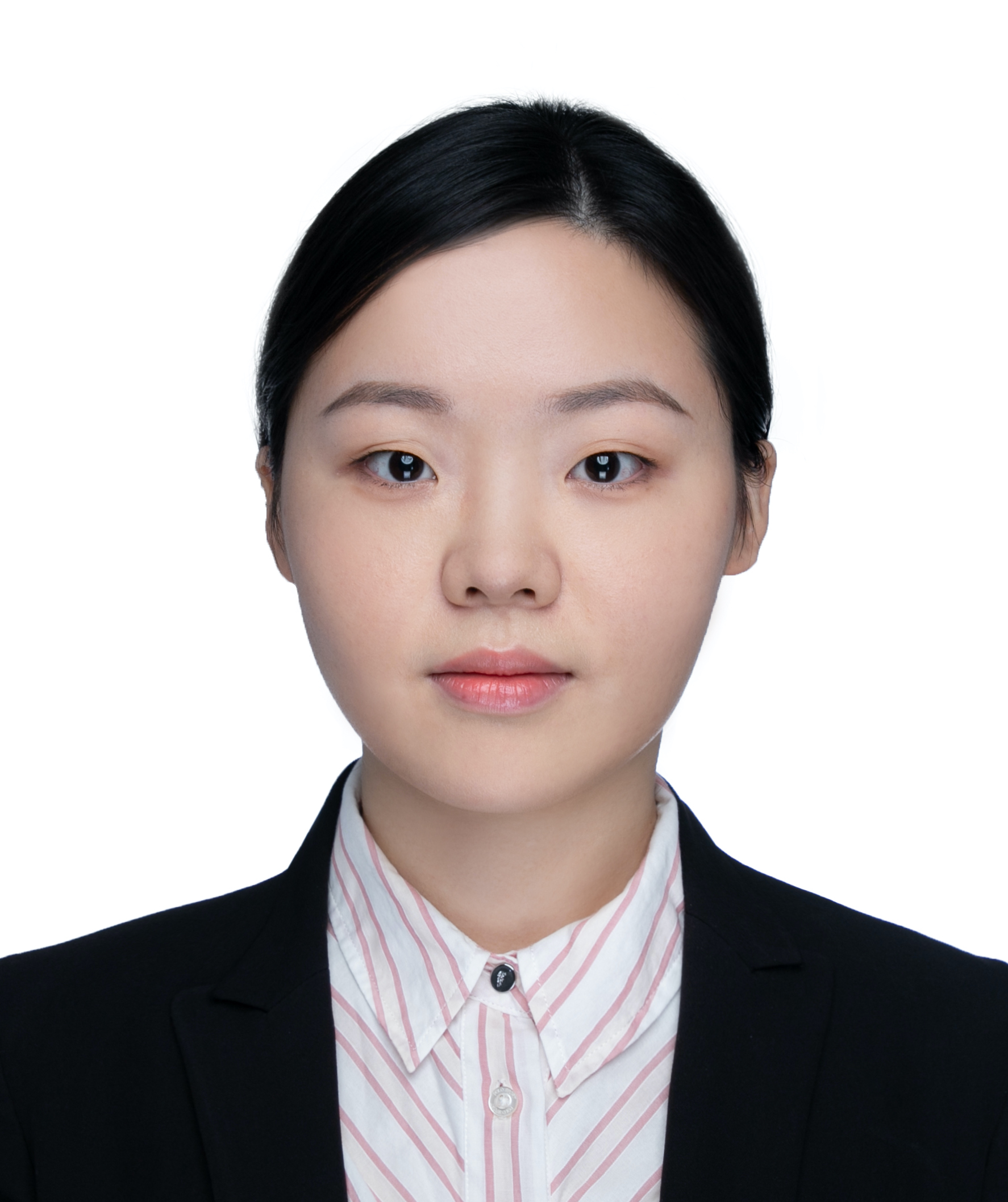}}]{Lingzhi Wang} is currently pursuing the Ph.D. degree with the Department of Systems Engineering and Engineering Management, The Chinese University of Hong Kong, Hong Kong. She received her bachelor degree from Harbin Institute of Technology in 2019. Her current research interests include recommender system, dialogue system and social media analysis. She has served as program committee member of NLP and AI top conferences like ACL and EMNLP.
\end{IEEEbiography}
\vfill

\begin{IEEEbiography}[{\includegraphics[width=1in,height=1.3in,clip,keepaspectratio]{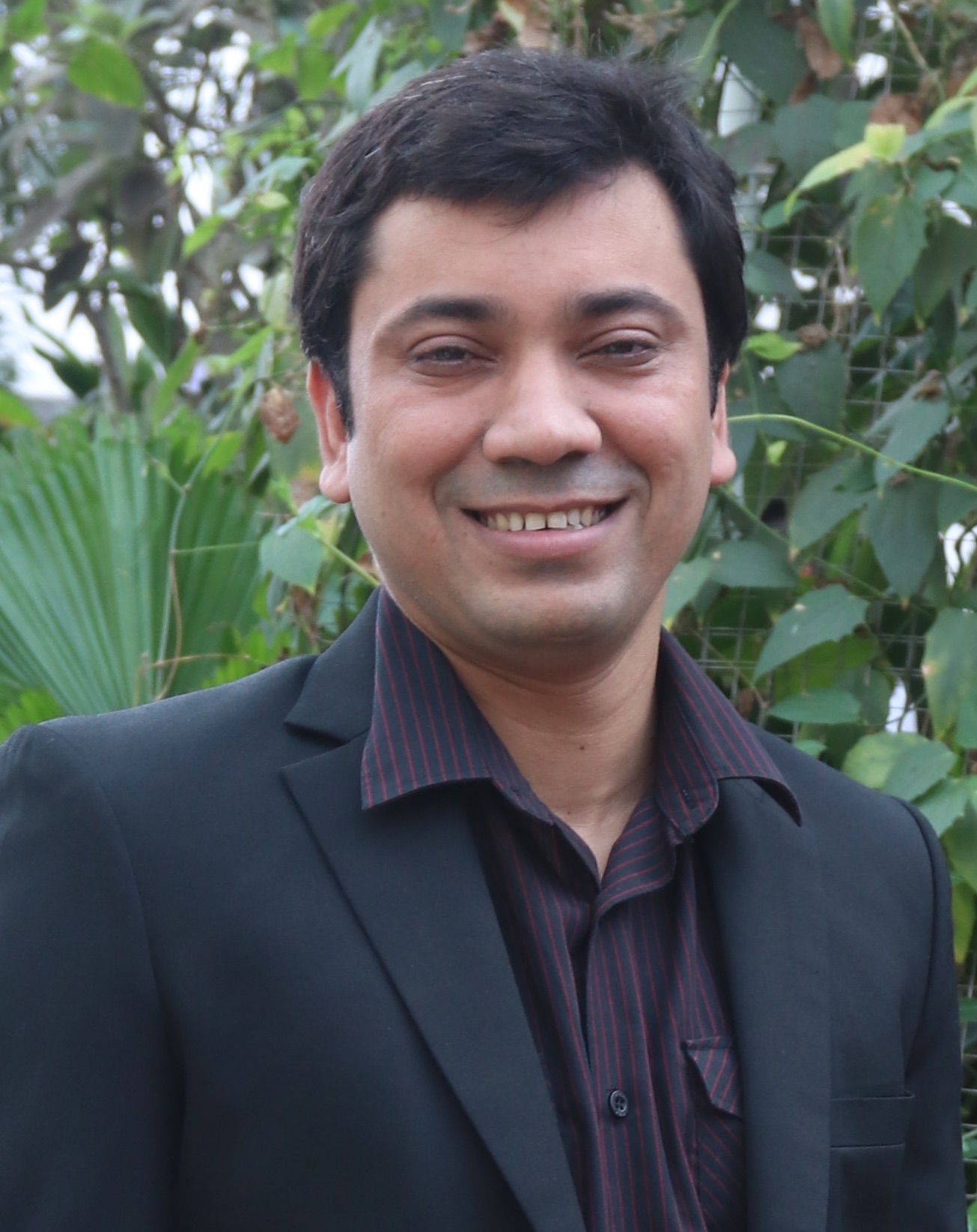}}]{Shafiq Joty} is currently an Associate Professor at the Nanyang Technological University, Singapore and a Senior Research Manager at Salesforce Research. He holds a PhD in {Computer Science from the University of British Columbia}. His work has primarily focused on developing language analysis tools and applications including question answering, text summarization and dialog systems. A significant part of his current research focuses on multilingual, multimodal NLP, and robustness of NLP models. He is a PC co-chair for SIGDIAL'23 and previously served as a senior AC for ACL’22 and EMNLP’21, and AC for ACL'19-21 EMNLP'19 and NAACL’21. He is an action editor for ACL-RR and was an associate editor for ACM TALLIP. 
His research contributed to 16 patents and more than 110 papers in top-tier NLP and ML conferences and journals.
\end{IEEEbiography}

% \newpage

\begin{IEEEbiography}[{\includegraphics[width=1in,height=1.25in,clip,keepaspectratio]{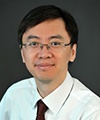}}]{Wei Gao}
is currently an Assistant Professor at Singapore Management University. Previously, he was a senior lecturer at Victoria University of Wellington, a Scientist at Qatar Computing Research Institute, and a Research Assistant Professor in the Chinese University of Hong Kong. His research interests intersect natural language processing, information retrieval, artificial intelligence and social computing. He broadly serves in the program committees of top conferences (e.g., ACL, EMNLP, SIGIR, IJCAI)
and reviews for the leading journals (e.g., ACM TOIS, IEEE TKDE) in his relevant field. He is currently an Associate Editor of ACM TALLIP and a member of standing review committees of TACL and CL.
\end{IEEEbiography}
% \vfill

\begin{IEEEbiography}[{\includegraphics[width=1in,height=1.25in,clip,keepaspectratio]{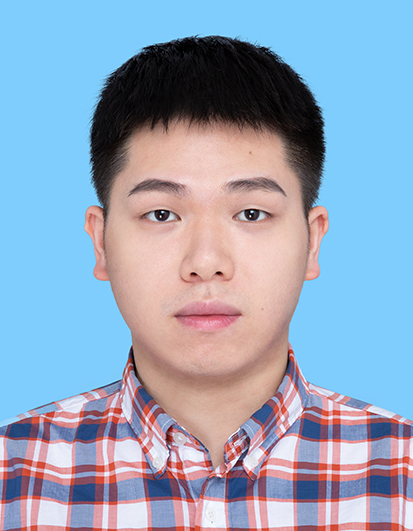}}]{Xingshan Zeng} is currently a researcher in Huawei Noah's Ark Lab. His research interests include natural language processing, social media analysis, recommender systems and speech translation. He got his PhD from the Department of Systems Engineering and Engineering Management, The Chinese University of Hong Kong in 2020. He has served as program committee member of NLP and AI top conferences like ACL, EMNLP, AAAI, IJCAI etc.
\end{IEEEbiography}
\begin{IEEEbiography}[{\includegraphics[width=1in,height=1.25in,clip,keepaspectratio]{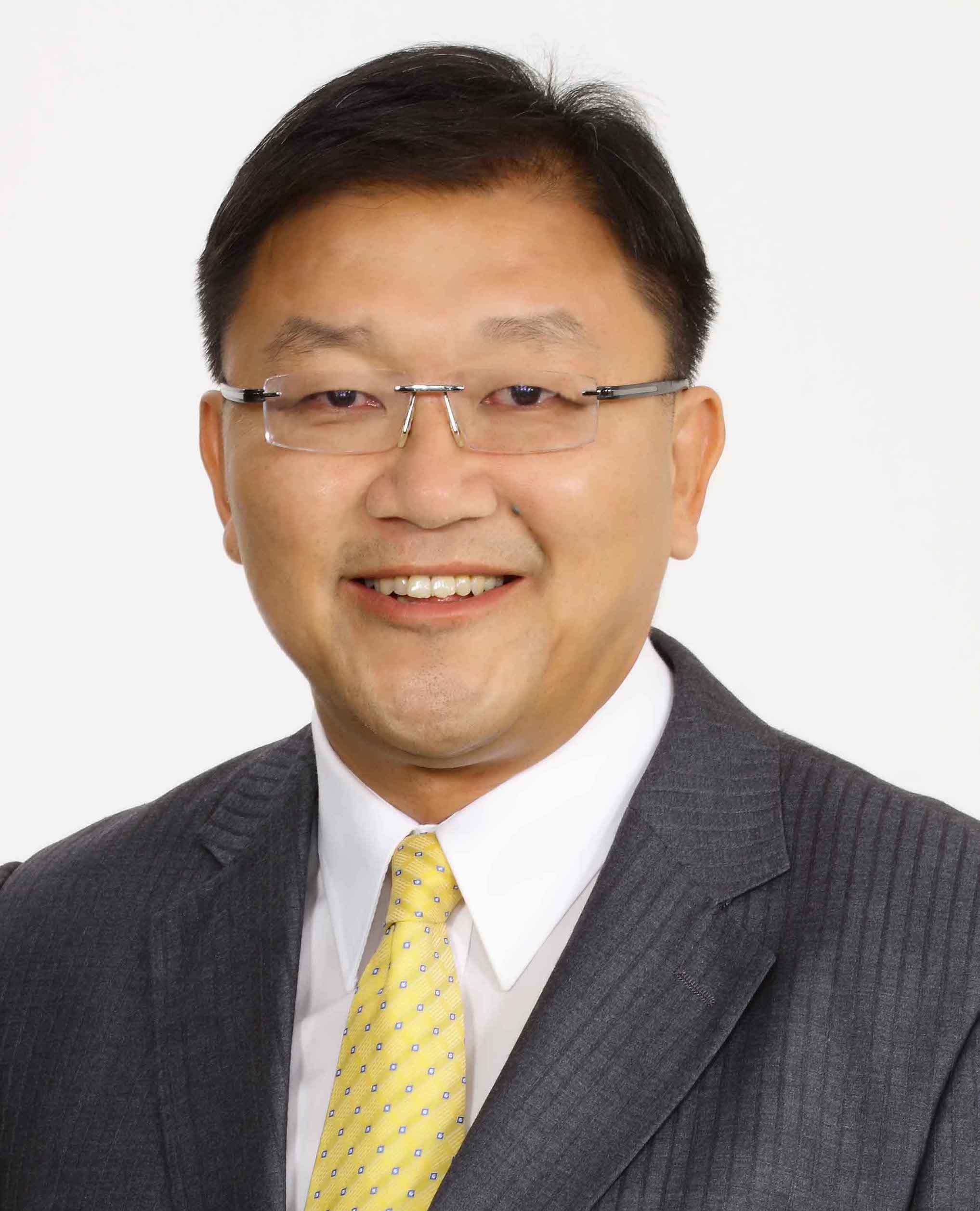}}]{Kam-Fai Wong} is currently a Professor in the Department of Systems Engineering and Engineering Management, The Chinese University of Hong Kong. He obtained his PhD from Edinburgh University in 1987. His research interest include Chinese computing, database and information retrieval. He is a fellow of ACL, and Senior Member of IEEE. He is the founding Editor-In-Chief of TALIP, and serves as associate editor of International Journal on CL. He is the Chair of Conference: General Chair of AACL2020, Co-Chair of NDBC2016, BigComp2016, NLPCC2015 and IJCNLP2011; the Finance Chair SIGMOD2007; and the PC Co-chair of lJCNLP2006.
\end{IEEEbiography}
\vfill

%\vfill


\end{document}